\documentclass[11pt,reqno,english]{amsart}
\usepackage[margin=0.95in]{geometry}

\usepackage[normalem]{ulem}

\usepackage[foot]{amsaddr}

\usepackage{
soul,
amssymb,
amsmath,
graphicx,
float,
tikz,
enumerate,
subcaption,
mathtools,
amsmath}

\usepackage[colorlinks,citecolor=red,urlcolor=blue,bookmarks=false,hypertexnames=true]{hyperref}

\newtheorem{theorem}{Theorem}[section]
\newtheorem{proposition}[theorem]{Proposition}

\newtheorem*{acknowledgements*}{Acknowledgements}

\theoremstyle{definition}

\newtheorem{remark}[theorem]{Remark}

\title{Understanding the geometry of deep learning with decision boundary volume}

\author{Matthew Burfitt}
\address[Matthew Burfitt]{Beijing Key Laboratory of Topological Statistics and Applications for Complex Systems, Beijing Institute of Mathematical Sciences and Applications (BIMSA)}
\author{Jacek Brodzki}
\address[Jacek Brodzki]{Mathematical Sciences, University of Southampton}
\author{Pawel D\l otko}
\address[Pawel D\l otko]{Dioscuri Centre for Topological Data Analysis, Institute of Mathematics, Polish Academy of Sciences}

\begin{document}

\begin{abstract}
    For classification tasks, the performance of a deep neural network is determined by the structure of its decision boundary, whose geometry directly affects essential properties of the model, including accuracy and robustness.
    Motivated by a classical tube formula due to Weyl, we introduce a method to measure the decision boundary of a neural network through local surface volumes, providing a theoretically justifiable and efficient measure enabling a geometric interpretation of the effectiveness of the model applicable to the high dimensional feature spaces considered in deep learning.
    
    A smaller surface volume is expected to correspond to lower model complexity and better generalisation. We verify, on a number of image processing tasks with convolutional architectures that decision boundary volume is inversely proportional to classification accuracy.  Meanwhile, the relationship between local surface volume and generalisation for fully connected architecture is observed to be less stable between tasks. Therefore, for network architectures suited to a particular data structure, we demonstrate that smoother decision boundaries lead to better performance, as our intuition would suggest. 
\end{abstract}

\maketitle

\section{Introduction}

    Deep neural networks have shown increasing success over the past several decades with various applications including computer vision \cite{Krizhevsky2017}, natural language processing \cite{Devlin2019}, and protein folding \cite{Senior2020}. However, due to a reliance on a huge number of trainable parameters obtained indirectly through a highly non-convex optimisation procedure, it is not easy to explain the surprising success of deep learning. As a consequence, effective generalisation to data embedded in high dimensional spaces \cite{Neyshabur2014, Zhang2017, Jiang2020} and robust strategies to circumvent network susceptibility to misidentification of phenomena such as adversarial perturbations \cite{Goodfellow2015} and out of distribution data \cite{Hendrycks2019} remain unresolved. Specifically, generalisation is the ability of a machine learning algorithm to correctly evaluate unseen data, making it one of the most important properties of any model. In particular, despite over-parametrisation, deep neural networks have been observed to generalise particularly well \cite{Neyshabur2018}.

    In classification tasks, the input space is divided into regions corresponding to label classes with the domain between different label regions being called the \emph{decision boundary}. In particular, the performance of a classifier is entirely determined by the structure of its decision boundary. Therefore, understanding the role that the geometry of decision boundaries plays in neural network generalisation provides a mathematical framework for further improving performance on challenging problems such as model selection \cite{Ramamurthy2019}, regularisation \cite{Bai2016, Chen2019, Liu2022}, and robustness \cite{Fawzi2018, Moosavi-Dezfooli2019}.

    Over the same period of time, topological data analysis has become a powerful tool for summarising geometric structure in data, a central method of which is persistent homology \cite{Carlsson09, Edelsbrunner2014}.
    However, it is well understood that persistent homology faces computational challenges when applied in a high-dimensional space, only effective in specific situations such as when the embedded structure of interest is sufficiently low-dimensional \cite{Chazal2016}.  
    The current computational limitations of studying significantly high dimensional ambient manifolds such as the decision boundaries of deep neural networks
    has renewed interest in developing new more powerful geometric methods that take advantage of the differential structure \cite{Bai2016, Zhang2022, Lei2020, Liu2022}.
    
\begin{figure}[ht!]
        \centering
         \includegraphics[width=0.5\textwidth]{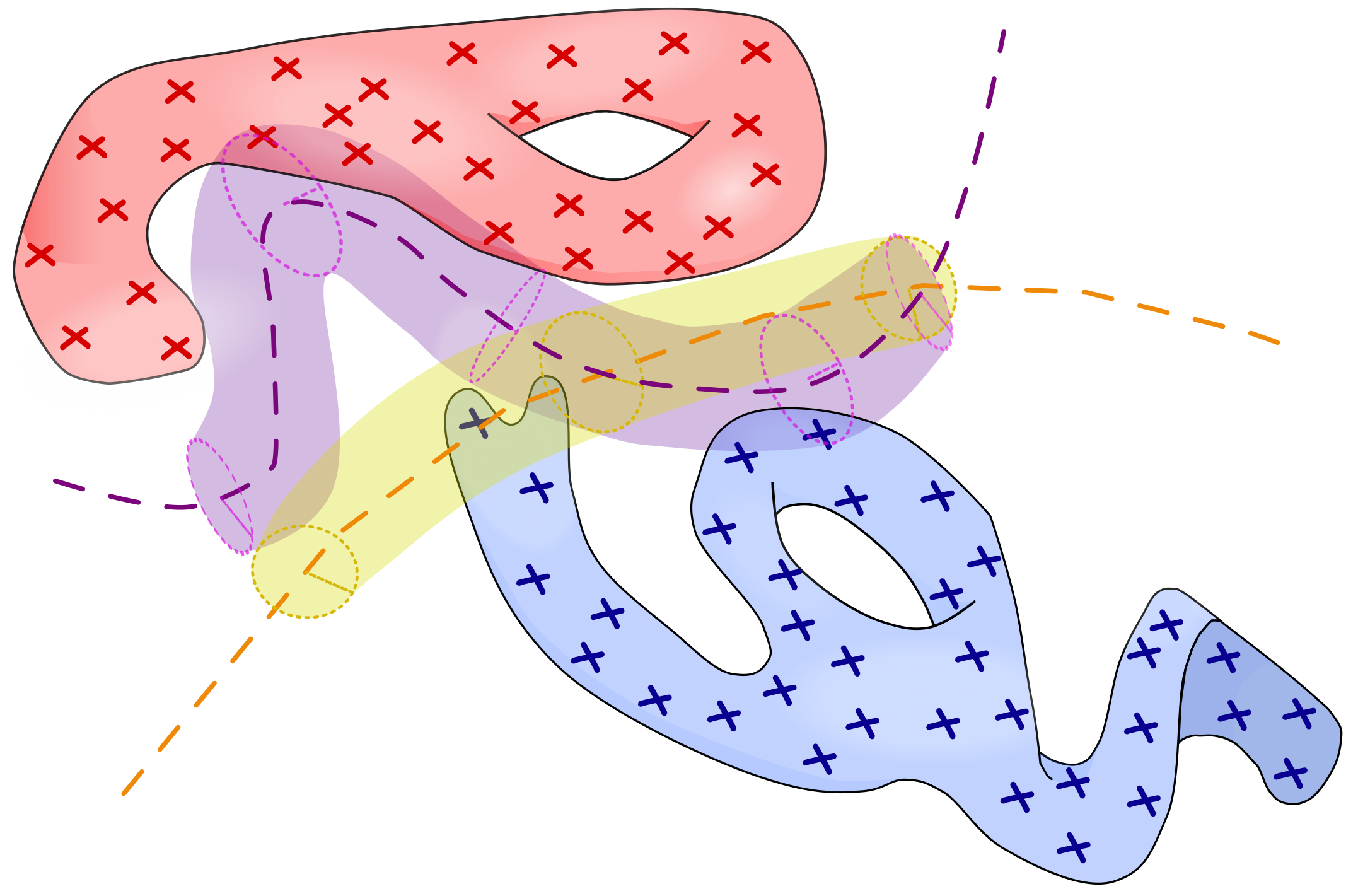}
        \caption{The volumes of a simpler decision boundary (orange) and its tubular neighborhood (yellow) should be smaller than the volumes of a more complex decision boundary and its tubular neighbourhood (purple) when measured in the vicinity of two data manifolds (red and blue).}
        \label{fig:TubularBoundary}
    \end{figure}

    In this paper, we focus on neural networks trained for classification. Previous studies have suggested that deep neural networks are biased towards learning simpler functions \cite{DePalma2019, Valle-Perez2019}.  Prior to evaluating labels, the function learned is in general continuous and smooth if the network activation functions are smooth. Therefore, it is reasonable to assume that a neural network approximates a smooth function in general. This would imply that the network decision boundaries form a finite union of codimension-$1$ submanifolds partitioning the networks decision regions, making them accessible to study from the perspective of differential geometry. A formula for the volume of the tubular neighbourhoods of a manifold embedded in a Euclidean space was first formulated by Weyl \cite{Weyl1939} in 1939, in terms of invariants dependent on the curvature of the manifold. The initial motivation for our work is that a tubular neighbourhood is a computationally feasible quantity in high dimensions, that for a small enough neighbourhood is directly proportional to the volume of the decision boundary of a deep neural network. In particular, we can use the size of the volume of a neural network decision boundary as a geometric measure of the functions complexity (see Figure~\ref{fig:TubularBoundary}).

    In lower dimensional settings, tube formulas have previously been successfully applied to obtain surface volume \cite{Edelsbrunner2014a} and to measure local curvature \cite{Chazal2017}. Therefore, measuring neighbourhood boundary volumes over a range of tube sizes also provides a framework in which measurements of neural network decision boundary curvature might be made in the future.

    To obtain boundary volume measures, we first identify regions of the decision boundary to measure then apply a Monte Carlo method combined with a procedure to generate adversarial examples. Three subspaces of the network input space are proposed: the space of all possible input vectors, a neighbourhood of the training data and a neighbourhood of points sampled near the decision boundary between training label classes. When combined, these measurements provide a geometric characterisation of the networks performance when compared to other similarly trained network functions.

    In the course of our work, we justify that it is precisely for reasons of concentration of measure on distances (see Section~\ref{sec:obsitcals} equation~\eqref{eq:ConstOfDist}) that conventional distance-dependent methods such as persistent homology generally fail to provide meaningful measurements in a high-dimensional setting. Specifically in the case of persistent homology, problems arising from the curse of dimensionality were studied in detail by Hiraoka, Imoto, Kanazawa and Liu \cite{Hiraoka2025}. Here, we demonstrate that the geometric simplifications which are a feature of high dimensions allow tubular neighbourhoods to become an efficient tool. In particular, due to the small neighbourhoods in which we make measurements, computationally efficient adversarial attack methods can be applied without compromising the accuracy of the results.

    \subsection{Main contributions} 
    
    Our main contributions are the following:
    \begin{itemize}
        \item 
            We introduce a new computational method to estimate the volume of a tubular neighbourhood of a neural network decision boundary, proposing three regions of the input space on which to make measurements. Moreover, by using a Monte Carlo method we obtain efficient volume measurements irrespective of the dimension.
        \item 
            Employing Weyl's tube formula and concentration of measure, we demonstrate that our proposed measurements provide quantities approximately directly proportional to the true surface volume of the decision boundary.
        \item
            Our experimental study with variations in dropout rate for the MNIST, Fashion MNIST, and CIFAR-10 data sets confirm our intuition that when the network architecture is well suited to the learning task, the better generalising networks have smaller decision boundaries.
        \item
            We also demonstrate that for fully connected architectures not suited to their task, boundary complexity can depend
            on the data set and optimisation procedure. However, it remains the case that over hyperparameter changes the optimal network performance coincides with critical points in the structure of the decision boundary volume.
    \end{itemize}

    \subsection{Related work} 

    In the case of model selection for lower dimensional data, Ramamurthy, Varshney, and Mody \cite{Ramamurthy2019} define a labeled \v{C}ech complex generalising methods commonly used in computational topology. The labeled \v{C}ech complex is applied with persistent homology to capture geometric features describing the region between label classes in which the decision boundary is contained. In particular, the features obtained can inform the choice of a model and generalisation performance. Chen, Ni, Bai, and Wang \cite{Chen2019} demonstrate a topological regularisation which penalises significant critical points occurring in the network function. Using Morse theoretic reasoning, this biases the network function to a simpler decision boundary topology. In the context of generalisation and adversarial robustness in deep learning, persistent homology has also been proposed as a complexity measure on the training data, aiding the choice of network architecture \cite{Carlsson2020}. Furthermore, persistent path homology has been applied to neural activation graphs \cite{Gebhart2019, Lacombe2021}, and a comparative persistent barcode has been developed and used to compare the connectivity of data representations in the final layer of a network throughout training \cite{Barannikov2022}. However, computational complexity provides a significant obstacle in each case. In addition, by restricting measurements to only the network weights and activations, the possibility of providing a direct geometric interpretation of the network function is lost.

    In an alternative direction, methods from differential geometry have been applied to construct a geometric regularisation procedure by Bai, Rosenberg, Wu, and Sclaroff \cite{Bai2016} using gradient flows targeting small local oscillation in the decision boundary to minimise the volume and curvature. Making use of differential geometric structure of the data manifold, \cite{Zhang2022} proposes normal adversarial risk by restricting to the perpendicular direction as a geometrically informed local component of adversarial risk. In another case, \cite{Lei2020} apply optimal transport on the data manifold to improve the efficiency of Generative Adversarial Networks (GANs), while an alternative method \cite{Yao2024} combines a manifold fitting procedure with GANs to improve performance. The differential geometry of neural network decision boundaries has also been considered from a theoretical perspective with a view toward generalisation in \cite{Liu2022}, leading the authors to propose a method for calculating the Euler characteristic of the decision boundary. Some direct empirical measurements of the curvature of decision boundaries have also been presented in \cite{Fawzi2018}.

    Other related work includes a measurement called boundary thickness that studies the average distance between level sets of two label classes of a network function using a Monte Carlo method \cite{Yang2020} and was shown to provide a measure of adversarial robustness. Considering the geometry of decision regions rather than the boundary, Cao and Gong \cite{Cao2017} measured the percentage decision region volumes for each class around benign and adversarial examples in order to distinguish between them, also making use of a Monte Carlo method.

\begin{acknowledgements*}
    The authors would like to thank John Harvey for his useful advice and discussion on parts of this work. JB was supported in part by the grant
    EPSRC EP/Y007484/1, Mathematical Foundations of AI: An Erlangen programme for Artificial Intelligence. PD is supported by the Dioscuri program initiated by the Max Planck Society, jointly managed with the National Science Centre (Poland),
    and mutually funded by the Polish Ministry of Science and Higher Education and the German Federal Ministry of Education and Research.
\end{acknowledgements*}

\section{Background}

    In this section we set out the notation used in the rest of the paper and introduce the necessary background on adversarial attacks relevant to developing neighbourhood boundary volume methods.
    We later make use of adversarial attacks to obtain an upper bound on the distance from a point to the network decision boundary.

    \subsection{Deep feed forward neural networks}

    Let $X$ denote a training data set consisting of a finite set of $n$-dimensional points lying in $[0,1]^n$ with corresponding discrete labels $Y$ taking values in the standard basis of $\mathbb{R}^m$, where $m$ is the number of classification classes. We let $N \colon \mathbb{R}^n \to \mathbb{R}^m$ be the function described by a feed forward neural network with $n$ inputs and $m$ outputs.
    The metric on the input space of network $N$ is denoted by $d\colon \mathbb{R}^n\times\mathbb{R}^n\to \mathbb {R}$. We recall that a feed forward neural network $N$ with $k$ layers of  sizes $n=l_0, l_1,\dots,l_k=m$ respectively is defined inductively by the composition of functions of the form
    \begin{equation}\label{eq:NN}
        \phi_i(W_ix+b_i)\colon{\mathbb{R}^{l_{i-1}}\to\mathbb{R}^{l_i}}
    \end{equation}
    where $i=1,\dots,k$, the $W_i$ are $l_{i-1}$ by $l_i$ weight matrices, the  $b_i\in \mathbb{R}^{l_i}$ are bias vectors, and $\phi_i\colon \mathbb{R}^{l_i} \to \mathbb{R}^{l_i}$ are non-linear continuous activation functions.
    Together, the $W_i$ and $b_i$ are the network's trainable parameters.
    In this work, activation functions $\phi_i$ are usually rectified linear functions $x_j \mapsto \max(0,x_j)$ in each dimension with the exception of the final layer where $\phi_k$ is a softmax function
    \[
        x_j \mapsto \frac{\exp{x_j}}{\sum_{s=1}^{m}\exp{x_s}}
    \]
    for $x=(x_1,\dots,x_m)$ a point in $\mathbb{R}^{m}$ and $j=1,\dots,m$.
    Layer $0$ of $N$ is called the input layer, layer $k$ the output layer, and the other layers are called hidden layers. A fully connected network is distinguished by having no restrictions on the entries of the trainable parameters during training. Even with a single hidden layer and under only mild assumptions, fully connected neural networks are universal function approximators \cite{Cybenko1989, Hornik1991}.

    A network $N$ is trained on labeled data set $(X,Y)$ through an optimisation procedure and associated cost function given as the mean value of loss functions $\mathcal{L}(N,x,y)\colon \mathbb{R}^n\times \mathbb{R}^m \to \mathbb{R}$ for each $x\in X$. There are standard choices of the loss function in the literature, in this work we apply the cross-entropy loss function along with the stochastic gradient descent (SGD) or Adam optimisers \cite{Kingma2015}.

    \subsection{Dropout regularisation}\label{sec:Drop}
    
    In the $i^{\text{th}}$ layer of a neural network $N$ for some $i=1,\dots,n$ and probability $p\in [0,1]$, dropout regularisation can be applied to the layer during training by removing each of the $\mathbb{R}^{l_i}$ coordinates with probability $p$.
    More precisely, at each training step take a random $l_{i-1}$ by $l_i$ matrix $D$ with entries in $\{0,1\}$ where $0$ appears with probability $p$ and $1$ with probability $1-p$.
    Then for the purposes of evaluating gradients, during the weight update optimisation procedure only, replace the network function definition given in equation~\eqref{eq:NN} by
    \begin{equation*}
        \phi_i\left( (D \cdot W_i)x+b_i \right)\colon{\mathbb{R}^{l_{i-1}}\to\mathbb{R}^{l_i}}
    \end{equation*}
    where $D \cdot W_i$ denotes the entry-wise multiplication of the matrices.
    To simplify the experimental setup, in this work we consider dropout only when applied to the final hidden layer of fully connected or convolutional neural networks.

    \subsection{Convolutional neural networks}
    
    Convolutional network layers are defined on data with the structure of a cubical array, such as an image. Here one applies homogeneously a number of trainable convolution operations to each local cubical patch in order to obtain entries of a next cubical layer, after which pooling operations are applied to reduce the number of entries \cite[\S~9]{Goodfellow2016}.
    \begin{figure}[ht!]
        \centering
         \includegraphics[width=0.975\textwidth]{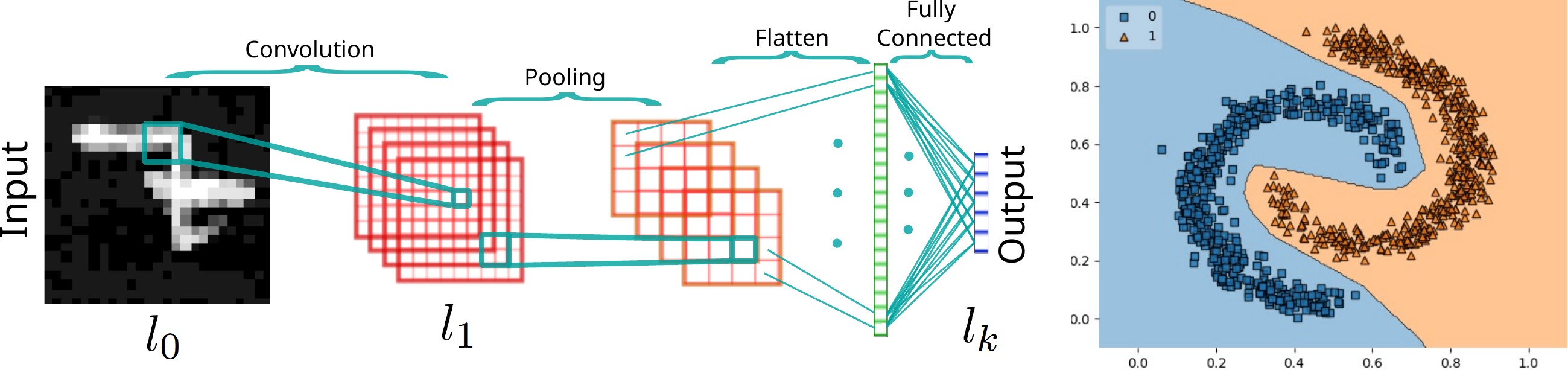}
        \caption{To the left, a demonstration of layer-wise operations in a convolutional neural network taking grayscale images as an input. On the right, a neural network trained to classify two classes of $2$-dimensional points. Class $0$ points are indicated by blue squares and the class $1$ points by orange triangles. The decision boundary is formed by the black line in-between the blue and orange decision regions of the two label classes.}
        \label{fig:ConvolutionalNetwork}
    \end{figure}
    At a chosen layer, flattening is implemented by forgetting the cubical structure and subsequent layers are obtained using a fully connected architecture until the output layer.
    However, even prior to flattening the convolution and pooling layers can be represented in the form of equation~\eqref{eq:NN}, though additional restrictions are placed on the layer parameters during training.

    \subsection{Decision boundaries}

    For each $x\in \mathbb{R}^n$, denote by $N_j(x)\in \mathbb{R}$ the output of $N$ in the coordinate $j = 1,\dots, m$.
    The prediction function
    \begin{equation*}
        p\colon \mathbb{R}^m\to \mathcal{P}(\{ 1,\dots,m \})
    \end{equation*}
    at $z=(z_1,\dots,z_m)\in \mathbb{R}^m$ is given by the set of $j\in \{1,\dots,m\}$ such that coordinates $z_j$ have maximal size, where $\mathcal{P}(S)$ is  the power set of $S$.
    The composition of $N$ with $p$ converts the continuous function $N$ to a function with a discrete image. The elements $1,\dots,m$ are identified with the standard basis vectors of $\mathbb{R}^m$ in $Y$ and are also referred to as \emph{labels} or \emph{class labels}. The preimages of $\{1\},\dots,\{m\}$ under $p \circ N$ are the \emph{decision regions} of each class label $1,\dots,m$, respectively.
    Note that for $x\in \mathbb{R}^n$, if $p(N(x))= \{i\}$ we abbreviate to $p(N(x)) = i$.
    
    The \emph{decision boundary} $B(N)\subseteq\mathbb{R}^n$ of $N$ is a set of points between decision regions of labels classified by $N$, that is
    \begin{align*}
        B(N) = \{ x \in \mathbb{R}^n \; | \; &\text{for every } \varepsilon>0 \text{ there are } y,z\in\mathbb{R}^n 
        \text{ with } \\ &  d(x,y) <\varepsilon,\: d(x,z)<\varepsilon \text{ and } p(N(y)) \neq p(N(z))  \}
    \end{align*}
    With smooth activation functions, the decision boundary of a feedforward neural network in a two-class classification problem is a manifold of codimension $1$ in $\mathbb{R}^n$ with probability $1$ when obtained using gradient descent from randomly selected initial weights.
    An example of such a decision boundary learned by a neural network is shown in Figure~\ref{fig:ConvolutionalNetwork}. 
    More generally, in a multiclass classification problem, the decision boundary is the union of the (non-empty) codimension $1$ manifold with boundary sitting between the decision regions of each pair of class labels.

    \subsection{Adversarial attack methods}\label{sec:FGSM}
    
    An \emph{adversarial example} at distance $\varepsilon > 0$ to a point $x\in \mathbb{R}^n$ (often chosen from the training set $X$) is an $a\in \mathbb{R}^n$ such that
    \begin{equation*}
        p(N(x)) \neq p(N(a)) \text{   and   } d(x, a) \leq \varepsilon.
    \end{equation*}
    Geometrically an adversarial example at $x$ is another point lying nearby $x$ on the other side of the decision boundary.
    An \emph{adversarial attack method} or \emph{adversarial attack} is a procedure for generating adversarial examples from a given trained network and point in its domain.
    For details on various adversarial attack methods see for example \cite{Xu2020, Yuan2019}.
    
    For this to work in practice, we consider only metric $d$ induced by the $l_{\infty}$-norm when generating adversarial examples.
    The fast gradient sign method (FGSM) \cite{Goodfellow2015} is a computationally efficient adversarial attack method obtained using the gradient of the loss function $\mathcal{L}$ locally at a point $x\in \mathbb{R}^n$ with a label $y$ by
    \begin{equation}\label{eq:FGSM}
        \text{FGSM}_{\varepsilon}(N,x,y) =
        x + \varepsilon\cdot\text{sign}(\nabla_{x}\mathcal{L}(N,x,y))
    \end{equation}
    where the function $\text{sign}$ rounds each non-zero coordinate to \textpm$1$ and $\nabla_{x}$ is the derivative with respect to $x$.
    Note that in contrast to the stochastic gradient descent method used to train the neural network, here we aim to maximise the value of the local loss function in order to obtain an adversarial example with a different label to the class of the original point $x$.

    Adversarial examples in this work are applied to determine an upper bound on the distance from a given point to the network decision boundary.
    There is a trade off between the effectiveness with which an adversarial attack finds an adversarial example, the computation time required to execute the attack, and the distance $\varepsilon$ over which the attack is effective.
    As we later make use of a Monte Carlo method, typically a large number of adversarial examples are generated.
    So our chosen adversarial attack methods should be computationally efficient.
    The  trade-off in accuracy resulting from our preference for efficiency is compensated for by the fact that adversarial examples are generated only at relatively short distance $\varepsilon$ than would generally be considered when evaluating network robustness for security reasons.
    For an example with networks trained on MNIST comparing the FGSM and a strong projected gradient descent adversarial attack over varying $\varepsilon$ values see \cite[Figure 3]{Raghunathan2018}.

    \begin{remark}\label{rmk:UseOfFGSM}
        In the experimental part of this work we apply our methods using the fast gradient sign adversarial attack.
        It should also be noted \cite{Andriushchenko2020} that the FGSM performs well with networks trained using rectified linear activation functions. That is, the linear nature of the adversarial attack method interacts effectively with the piecewise linear network function.
        As a consequence, when using the methods set out in the paper with the FGSM we can expect to provide a more reliable bound for a network with rectified linear activation functions.
        
        We also observed that with networks trained over many epochs beyond the point of optimal test loss, points not located closely to the decision boundary were increasingly likely to have vanishingly small gradient values $\nabla_{x}\mathcal{L}(N,x,y)$, with many being computed as zero in practice.
        In particular, the vanishingly small gradient can be explained by high convergence of the network function in regions of the input space far from the decision boundary.
        
        Consequently, the vanishingly small gradients resulted in the FGSM being significantly less effective for the purposes of measuring the decision boundary of a neural network in the vicinity of points not specifically selected within a reasonable proximity of the training data.
        Therefore, if it were desirable to perform experiments in these regions, it would be advisable to make use of a different adversarial attack method.
    \end{remark}

\section{Obstacles to quantifying the geometry of decision boundaries in high dimensions}\label{sec:obsitcals}

    Before proposing a new procedure to measure neural network functions with higher dimensional inputs, we first discuss some of the consequences of the unintuitive geometry that occurs in the structure of high dimensional Euclidean spaces. Since typically in deep learning the dimension of the feature space is very large, it is important to understand where conventional intuition breaks down and conversely of what simplifications in structure we might be able to take advantage.

    The curse of dimensionality is often used to refer to a number of mathematical phenomena that make studying data with a large number of parameters difficult \cite{Gorban2018}.
    Perhaps the easiest example of the curse of dimensionality is for grid samples.
    In this case, a $100$ by $100$ grid in $2$ dimensions requires $10000$ points to be sampled.
    For an arbitrary dimension $n$, to achieve the same precision of coverage we would require $100^n$ points to be sampled, a quantity that quickly becomes impractical as $n$ grows.
    
    Many properties presenting a curse of dimensionality can be expressed in terms of a concentration of measure around a certain set as dimensions increases. 
    An example of this is given in \cite{Gorban2016}, where it is shown that the number $N$ of randomly selected vectors in $\mathbb{R}^m$ that are almost certainly orthogonal grows exponentially with $n$.
    More precisely, we have the following proposition.
    Given $\varepsilon>0$, say two vectors are $\varepsilon$-orthogonal if the dot product of their unit vectors is less than $\varepsilon$.
    \begin{proposition}[\cite{Gorban2016}]\label{prop:NAlmostOthonomal}
        For $\varepsilon,\theta>0$, a set of $N$ uniform random vectors chosen in the unit ball in $\mathbb{R}^n$ are pairwise $\varepsilon$-orthogonal with probability greater than $\theta$ when
        \begin{equation*}\label{eq:NAlmostOthonomal}
            N \leq e^{\frac{\varepsilon^2n}{4}}\sqrt{\log(1/\theta)}.
        \end{equation*}
    \end{proposition}
    The above bound can be derived from the property that as $n$ increases the surface volume of an $n$-sphere is concentrated in a smaller neighbourhood of its equator.
    Such concentration of measure properties in high dimensional spaces are studied and applied throughout the literature \cite{Ledoux2001,Talagrand2001, GORBAN2016b, GORBAN2017, Tyukin2018},
    including the more general waist concentration results due to Gromov \cite{Gromov2009}.
    
    A manifestation of the curse of dimensionality that is perhaps most relevant to measuring the geometry of network decision boundaries is the concentration of distance in high dimensional $l_p$-spaces.
    Denote by $\| - \|_p$ the $p$-norm on $\mathbb{R}^n$. 
    For independent randomly sampled points $x^n_1,\dots,x^n_m$ drawn from $[0,1]^n$ we have that
    \begin{align}\label{eq:ConstOfDist}
        \text{if}& \;\;\;
        \lim_{n\to \infty}var\left( \frac{\| x^n_i \|_p}{\mathbb{E}\| x^n_i \|_p} \right) = 0 \nonumber
        \\
        \;\text{then}&\;\;\;
        \lim_{n\to \infty}\frac{\max_{i,j}\| x^n_i-x^n_j \|_p-\min_{i,j}\| x_i^n-x^n_j \|_p}{\min_{i,j}\| x^n_i-x^n_j \|_p} = 0
    \end{align}
    where $i,j=1,\dots,m$.
    Equation~\eqref{eq:ConstOfDist} is a special case of a theorem of Beyer, Goldstein, Amakrishnan, and Shaft \cite{Beyer99}.
    The result can be interpreted as saying that in high dimensions most points have a similar pairwise distance.
    
\begin{figure}
        \includegraphics[width=0.5\columnwidth]{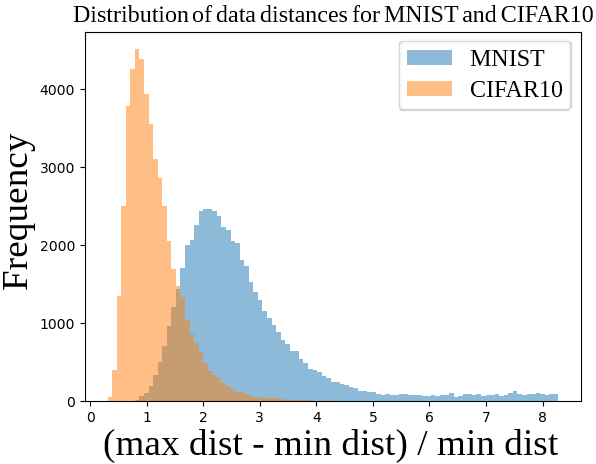}
        \caption{Frequency of $\frac{\max_{j}\| x-x_j \|_2-\min_{i,j}\| x-x_j \|_2}{\min_{j}\| x-x_j \|_2}$ values for data points $x \neq x_j$ in the MNIST and CIFAR-10 data sets, respectively. In particular, the width of the distribution for the higher dimensional CIFAR-10 data is smaller than width of the MNIST data distribution.}
        \label{fig:DistFrequencyComparison}
    \end{figure}

    The effect of Equation~\eqref{eq:ConstOfDist} on the image data considered later in this work is easily demonstrated. For example, the CIFAR-10 training set contains $50000$ images of dimension $3072$. Taking $x_i,\: x_j$ to lie in the CIFAR-10 training set, evaluating the expression inside the limit of equation~\eqref{eq:ConstOfDist} yields an approximate value of $10.056$. Compared to the maximal distance of $\sqrt{3072}\approx 55.426$
    in the space of possible data points within the $n$-cube, $10.056$ is already relatively small. Furthermore, a frequency plot of the distribution of values in equation~\eqref{eq:ConstOfDist} for fixed $i$ indices in the cases of the MNIST and CIFAR-10 data sets using the Euclidean norm can be seen in Figure~\ref{fig:DistFrequencyComparison}. In particular, the distribution of values for CIFAR-10 is seen to be significantly narrower than that for MNIST.

    An alternative perspective is that while our intuition from low dimensions no longer applies, the geometry of high-dimensional data is far less variable, as demonstrated in Proposition~\ref{prop:NAlmostOthonomal} and equation~\eqref{eq:ConstOfDist}. This phenomenon is known as the Blessing of Dimensionality \cite{Donoho2000, Anderson2014}. Techniques applicable to high-dimensional structures might therefore be designed to exploit these simplifications in the high dimensional geometry of Euclidean space. 
  
    One of the main contributions of this work is to demonstrate a method suitable for large deep neural networks that can explore the shape of the network function by describing the size of the boundary within a high dimensional setting.
    This measurement is then targeted at particular regions of the input space and shown to provide useful information about the effectiveness of the network function.

\section{Tube formulas and decision boundaries}

In this section we discuss how measurements of the neighbourhood of a neural network decision boundary are related to the geometry of a tubular neighbourhood. We archive this through a comparison to the existing mathematical literature on tubular neighbourhoods of embedded closed manifolds. More precisely, our main object of interest is the volume of the decision boundary of a deep neural network, and we justify that for a small enough neighbourhood parameter $\varepsilon$, the neighbourhood boundary volume is approximately proportional to the true volume of the decision boundary of the network. Therefore, by measuring a neighbourhood of a neural networks decision boundary, we gain useful geometric information about the effectiveness of the network function.

\subsection{Weyl's tube formula}\label{sec:TubeFormula}
   
    The volume $V^{\mathbb{R}^n}_M$  of an $\varepsilon$-tubular neighbourhood of a compact $q$-dimensional manifold $M$ embedded in $\mathbb{R}^n$ (with respect to the $l_2$ metric) was studied in a classical paper of Weyl \cite{Weyl1939} in 1939 and has since inspired developments in differential geometry \cite{Gray2004}.
    Weyl's tube formula is a polynomial in the tube radius $\varepsilon$ with coefficients containing invariants $k_{2i}(M)$ dependent on the curvature of $M$.
    The full expression of Weyl's tube formula is given by
    \begin{equation}\label{eq:TubeFormula}
        V^{\mathbb{R}^n}_M(\varepsilon) = 
        \frac{{(\pi\varepsilon^2)}^{(n-q)/2}}{((n-q)/2)!}
        \sum^{[q/2]}_{i=0}
        \frac{\varepsilon^{2i}k_{2i}(M)}{(n-q+2)(n-q+4)\cdots(n-q+2i)}
    \end{equation}
    for $\varepsilon$ small enough so that the $\varepsilon$-tubular neighbourhood does not self intersect.
    In particular, the formula is valid for closed manifolds and manifolds with boundary or corners.
    Moreover, the invariant $k_{0}(M)$ is the volume of the manifold $M$. When $M$ is closed and even dimensional,  the Gauss-Bonnet theorem may be applied to realise $k_{q/2}(M)$ as $(2\pi)^{q/2}$ times the Euler characteristic of $M$.

\subsection{Measuring the volume of neural network decision boundaries}\label{sec:MeasuringBoundaries}

    In this work, we are interested in measuring the $\varepsilon$-neighbourhood volume of the decision boundaries of a deep feedforward neural network.
    Therefore, we are primarily interested in the co-dimension $1$ case of equation~\eqref{eq:TubeFormula}, when $q=n-1$.
    We propose $\varepsilon$-neighbourhood volume because it is a quantity that might be viably measured using Monte Carlo methods even in high dimensions, and it is a useful proxy for the volume of the decision boundary, which we can interpret rigorously using Equation~\eqref{eq:TubeFormula}.
    
    Specifically, the decision boundary of a feedforward neural network used for classification in a two class classification problem trained with gradient descent is a manifold of codimension $1$ in $\mathbb{R}^n$ with probability almost 1.
    Moreover, if the network has smooth activation functions such as $\tanh$, the network decision boundary is a smooth manifold.
    In this case, by setting $q=n-1$ equation~\eqref{eq:TubeFormula} shows immediately that the volume of an $\varepsilon$-neighbourhood of the decision boundary is approximately proportional to the volume of the decision boundary, providing $\varepsilon>0$ is small. More precisely, since equation~\eqref{eq:TubeFormula} is a polynomial in $\varepsilon$ with the lowest degree term being linear in $\varepsilon$, for small $\varepsilon$ values $V^{\mathbb{R}^n}_M$ is approximately linear in a scalar multiple of $k_{0}(M)$, the volume of $M$.
    
    When the activation functions of the network are piecewise linear, such as for rectified linear activation functions, then the decision boundary is a piecewise linear manifold.
    In this case we would expect the network function to be approximating a smooth manifold. However, obtaining an explicit tubular formula in this situation turns out to be a more difficult problem than for the smooth case \cite{Cheeger86, Federer2014}.
    The fact that any piecewise linear manifold can be arbitrarily closely approximated by a $k$-differentiable manifold is a consequence of a classical result of Whitehead \cite[Theorem~9]{Whitehead1940}.
    
    For a multiclass classification problem the decision boundary consists of a union of manifolds with boundary, each lying between pairs of class decision regions.
    In this case, any intersection between the manifolds will with almost certain probability coincide with one of the manifold boundaries, which is a codimension-$2$ manifold.
    Provided $n$ is large and $\varepsilon$ is small enough, the effect of these intersections on the neighbourhood boundary volume should be negligible in comparison to the total volume of the decision boundary.
    This can be empirically verified by making use of the procedure presented later in this work in combination with adversarial attacks targeted at specific classification regions.

\section{Justification for the use of $l_\infty$ adversarial measurements}

    Accurate and efficient measurements of distances to the decision boundary are required to provide useful information about the structure of a network function. In Section~\ref{sec:FGSM} we described the FGSM adversarial attack, which makes use of the $l_{\infty}$ norm rather than the $l_2$ (Euclidean norm) for measurements of the distance to the boundary. Meanwhile, volume measurement using Weyl's tube formula discussed in Section~\ref{sec:TubeFormula} depends on the Euclidean metric.
    A simple way to circumvent this potential problem would be to use an $l_2$ adversarial attack method.
    However, for reasons of computational efficiency, it is preferable to make use of the FGSM.
    We justify this choice using Theorem~\ref{thm:RandomPlaneDistance}.
    Moreover, we conclude that the results of applying the methods presented later in this work using an $l_\infty$ adversarial attack are only affected up to multiplication by a scalar.
    In particular, scaling our outputs does not affect the method, as in Section~\ref{sec:TubeFormula} we assume the output of our boundary volume measurements is only directly proportional to the true volume of the decision boundary.  

    Denote Euler's gamma function by $\Gamma \colon \mathbb{C} \to \mathbb{C}$; this logarithmically convex continuous extension of the factorial function is important in our context as it appears in the general volume formula for an $n$-dimensional sphere.
    Let $d_2$ be the metric obtained from the $l_2$ norm and $d_\infty$ the metric obtained from the $l_\infty$ norm.
    Suppose $U$ is a subset of $\mathbb{R}^n$ and $x\in \mathbb{R}^n$, then for any metric $d$ on $\mathbb{R}^n$ define
    \[
        d(x,U) = \min_{u\in U}d(x,u).
    \]
    \begin{theorem}\label{thm:RandomPlaneDistance}
        Let $U$ be a randomly selected hyperplane in $\mathbb{R}^n$ and $x$
        a point in $\mathbb{R}^n \setminus U$, then
        \begin{enumerate}
            \item
                $\mathbb{E}\left(\frac{d_2(x,U)}{d_\infty(x,U)}\right) =
                \frac{\sqrt{n}}{\sqrt{\pi}}
                \frac{\Gamma(\frac{n}{2})}{\Gamma(\frac{n+1}{2})}$
                and
            \item 
                $\text{var}\left(\frac{d_2(x,U)}{d_\infty(x,U)}\right)\leq\frac{\pi - 2}{n}$.
        \end{enumerate}
    \end{theorem}
    
    The most important consequences of Theorem~\ref{thm:RandomPlaneDistance} are the following.
    Firstly, part $(1)$ implies $\mathbb{E}\left(d_2(x,U) / d_\infty(x,U)\right)$ approaches $\sqrt{\frac{2}{\pi}}$ from above as $n$ goes to infinity. Secondly, by applying Chebyshev's inequality to part (2) we obtain the measure concentration inequality
    \begin{equation}\label{eq:ConsetrationOfNorm}
        P\left(\left| \frac{d_2(x,U)}{d_\infty(x,U)} - \mathbb{E}\left(\frac{d_2(x,U)}{d_\infty(x,U)}\right) \right| \geq \varepsilon \right) \leq \frac{\pi - 2}{n \varepsilon^2}.
    \end{equation}
    A proof of Theorem~\ref{thm:RandomPlaneDistance} can be found in Appendix~\ref{sec:Proof}.
    
    When considered in the context of the sizes of dimensions arising in image classification tasks used in the experiential part of this work, the bound in equation~\eqref{eq:ConsetrationOfNorm} is already meaningful.
    More precisely, if $n=3000$ (approximately the dimension of the CIFAR-10 data set) and $\varepsilon = 0.05$, then the bound from equation~\eqref{eq:ConsetrationOfNorm} is $0.152$.
    However, the application of Chebyshev's inequality is in general far from optimal,
    as can be verified in our case with numerical simulations.
    For example, an empirical estimation of the deviation from the expected value on the left-hand side of equation~\eqref{eq:ConsetrationOfNorm}
    when $n=3000$ and $\varepsilon = 0.05$ using $10^4$ samples is already effectively zero for the purposes of practical measurements. 
    
    In the case of neural network decision boundaries, it has been empirically demonstrated \cite{Fawzi2018} that the decision boundaries at an adversarial example in the vicinity of data points are flat in almost all directions.
    Therefore, the optimal $l_2$ and $l_\infty$ adversarial examples should in general lie within the same high dimensional low curvature submanifold.
    Which means that Theorem~\ref{thm:RandomPlaneDistance} can also be applied to our setting to conclude that the difference between $l_2$ and $l_\infty$ measurements is approximately multiplication by a scalar.

\section{Methodology}\label{sec:methodology}

    In order to obtain useful information about a neural network function with regards to qualities such as generalisation, we require an effective measurement of the decision boundary volume.
    In this section we set out our procedure for measuring the neighbourhood boundary volume of a neural network function.
    This is carried out in the following steps. 
    First, we select an adversarial attack method; the effectiveness of this method within a close vicinity of training data will determine the accuracy of the volume estimate.
    Secondly, we select a region of the input data space on which to measure the neighbourhood boundary volume; our selection aims to reveal the complexity of the function on different scales.
    Finally, to achieve effective volume estimation even in high dimensions, we measure the neighbourhood boundary volume of a neural network function using a Monte Carlo method combined with the adversarial attack procedure. In particular, this method can straightforwardly provide measurements at a range of $\varepsilon$ values.
    
    Three regions on which measurements might be made are provided.
    We denote the resulting values by $\text{\bf Bvol}$, a measurement with respect to possible inputs; $\text{\bf TrainBvol}$, a measurement around the training data; and $\text{\bf LAdvBvol}$, a measurement between training label classes.
    The latter two proposed measurements also depend on an additional $\delta$ variable to determine their local search radius.
    
    We conclude the section by discussing some technicalities in the practical implementation of the introduced boundary volume measures and why they do not cause problems in the high dimensional setting.
    In the subsequent experimental part of this work, we restrict ourselves to using the FGSM as the adversarial attack method.

    \subsection{Estimating $\varepsilon$-neighbourhood boundary volume}\label{sec:BndVol}

    Given a region $\mathcal{U}\subseteq \mathbb{R}^n$, $\varepsilon > 0$ and an adversarial attack method $A$ for obtaining adversarial examples at a distance as near as possible to the decision boundary, we propose to obtain a bound on the volume of the $\varepsilon$-neighbourhood of the decision boundary of network function $N$ in $\mathcal{U}$ through Monte Carlo sampling of $\mathcal{U}$.
    As a point is determined to be within $\varepsilon$ of the decision boundary only when an adversarial example less than a distance of $\varepsilon$ away can be obtained, our volume measurement is in principle a lower bound on the true $\varepsilon$-neighbourhood boundary volume. In particular, the quality of the estimated bound depends on the reliability of the adversarial attack method $A$ and the Monte Carlo error.
    For the purposes of the theory discussed in this section, given $x \in \mathbb{R}^n$ denote by $A(x)$ the nearest adversarial example to $x$ determined by adversarial attack $A$ in terms of Euclidean distance.
    
    By applying the law of large numbers, Monte Carlo methods can be used to estimate almost any quantity that can be expressed probabilistically.
    In the case of volume, given a large uniformly distributed sample of a subspace $\mathcal{U}$ of $\mathbb{R}^n$, we obtain an estimate of the volume of a sub-region $\mathcal{V} \subseteq \mathcal{U}$ by counting the proportion of points that fall inside $\mathcal{V}$.
    In the present work, $\mathcal{V}$ is the $\varepsilon$-neighbourhood of the decision boundary of a neural network function $N$ intersected with a chosen region $\mathcal{U}\subseteq \mathbb{R}^n$.
    
    More precisely, in our case, the $\varepsilon$-neighbourhood boundary volume $\text{Bvol}_\varepsilon(N,\mathcal{U})$ is expressed probabilistically by 
    \begin{equation}\label{eq:Bvol}
        \text{Bvol}_\varepsilon(N,\mathcal{U})
        = \text{Volume}(\mathcal{U})\cdot P \left( d(x,A(x)) \leq \varepsilon \; | \; x\in \mathcal{U} \right).
    \end{equation}
    In practice and in all our experiments in the next section, it is convenient to set $\text{Volume}(\mathcal{U}) = 1$ as the remainder of the right hand side of equation~\eqref{eq:Bvol} will be straightforward to compute. Furthermore, as detailed in Section~\ref{sec:MeasuringBoundaries}, it is only necessary to obtain volume measurements up to multiplication by the same scalar value.
    
    The quantity $\text{Bvol}_\varepsilon(N,\mathcal{U})$ can be estimated using a Monte Carlo method by computing the proportion of points satisfying $d(x,A(x)) \leq \varepsilon$ from a large uniform random sample of points in $\mathcal{U}$, with some degree of error depending on the sample size.
    More explicitly, the Monte Carlo method is applied to estimate the Bernoulli parameter $\text{Bvol}_\varepsilon(N,\mathcal{U})$ of the random variable
    \begin{equation*}
        X(x) =
        \begin{cases}
            1 & \text{if} \: x \in d(x,A(x)) \leq \varepsilon \\
            0 & \text{otherwise.}
        \end{cases}
    \end{equation*}
    over uniformly distributed $x\in \mathcal{U}$. Therefore, denoting by $\widehat{\text{Bvol}}^l_\varepsilon(N,\mathcal{U}))$ the mean estimator of $l$ independent identically distributed copies of $X$ and applying the central limit theorem, we obtain
    \begin{equation}\label{eq:MCerrorBound}
        P\left(\left|\text{Bvol}_\varepsilon(N,\mathcal{U})-\widehat{\text{Bvol}}^l_\varepsilon(N,\mathcal{U})\right| \leq \zeta\right)
        \to
        \Phi\left(\zeta\right)
        \sqrt{\frac{\text{Bvol}_\varepsilon(N,\mathcal{U})(1-\text{Bvol}_\varepsilon(N,\mathcal{U}))}{l}}
    \end{equation}
    as $l\to \infty$, where $\Phi$ is the cumulative distribution function of the standard normal distribution. Alternatively, an application of standard results on concentration of measure can be used to obtain a formula for Monte Carlo errors \cite[\S 3]{Tang2024}, in our context resulting in
    \begin{equation}\label{eq:MCerrorBound2}
        P\left(
        \left|\text{Bvol}_\varepsilon(N,\mathcal{U})-\widehat{\text{Bvol}}^l_\varepsilon(N,\mathcal{U})\right|
        \geq \xi\right) 
        \leq
        e^{-2m\xi^2/\text{Volume}(\mathcal{U})^2}.
    \end{equation}
    Importantly, both equations~\eqref{eq:MCerrorBound}~and~\eqref{eq:MCerrorBound2} do not explicitly depend on the dimension, providing a large advantage over conventional methods of integration when applied to estimating volumes in high dimensions.
    
    As discussed in Section~\ref{sec:MeasuringBoundaries}, our aim is to use the $\varepsilon$-neighbourhood boundary volume as an alternative to the impractical measurement of the boundary volume alone, and for this reason we would like to choose $\varepsilon$ as small as possible.
    However, the choice of $\varepsilon$ must also be large enough to ensure that the Monte Carlo estimation is applied to a substantial enough fraction of $\mathcal{U}$ to give a reliable result.
    This conflict creates a trade-off in choice of $\varepsilon$ size in practice.

    \subsection{\text{\bf Bvol}}\label{sec:Bvol}

    Since we assume all training data lies in $[0,1]^n$, the most straightforward region to choose as the sample space would be $\mathcal{U}=[0,1]^n$.
    This choice corresponds to measuring the volume of an $\varepsilon$-neighbourhood of the decision boundary on the entire space of possible data points.
    We use the shorthand
    \begin{equation*}
        \text{\bf Bvol} = \text{Bvol}_\varepsilon(N, [0,1]^n)
    \end{equation*}
    to denote the corresponding $\varepsilon$-neighbourhood boundary volume measurement.
    However, as the training data is usually assumed to be sampled from a much lower dimensional manifold contained in a small region of $[0,1]^n$, measuring $\varepsilon$-neighbourhood boundary volume on the whole of $[0,1]^n$ captures structure of the decision boundary unrelated to the classification task.
    Moreover, the practical application of the procedure with the FGSM may also suffer from the problem of vanishing gradients highlighted in Remark~\ref{rmk:UseOfFGSM}, again due to the fact that most sample points need not lie in the vicinity of the data manifold.
    
    We therefore propose two additional sample spaces $\mathcal{U}$ on which to measure the $\varepsilon$-neighbourhood boundary volume.
    The first, $\text{\bf TrainBvol}$, measures the global complexity of the decision boundary restricted within a region relevant to training data.
    The second, $\text{\bf LAdvBvol}$, analyses the local complexity of the decision boundary between classification classes.

    \subsection{\text{\bf TrainBvol}}\label{sec:TrianBvol}
    
        Given training data $X$ and $\delta,\: \varepsilon > 0$, set
        \begin{equation*}
            \mathcal{U} = \bigcup_{x\in X} B_{\delta}(x)
        \end{equation*}
        where $B_\delta(x) = \{ x'\in \mathbb{R}^n \; | \; d_\infty(x,x') \leq \delta \}$ is the ball of radius $\delta$ around $x$.
        We use the shorthand
        \begin{equation}\label{eq:NearDef}
            \text{\bf TrainBvol} = \text{Bvol}_\varepsilon(N, \cup_{x\in X}B_{\delta}(x))
        \end{equation}
        to denote the corresponding $\varepsilon$-neighbourhood boundary volume measurement.

        We note that {\bf TrainBvol} depends on the parameters $\varepsilon$ and $\delta$.
        These parameters need to be selected in a sensible way to obtain a meaningful measurement.
        The selection of $\varepsilon$ and $\delta$ is discussed prior to the experimental results at the beginning of Section~\ref{sec:Experiments}.
        
\begin{figure}[ht!]
            \centering
            \def\svgwidth{200pt}
\begingroup%
  \makeatletter%
  \providecommand\color[2][]{%
    \errmessage{(Inkscape) Color is used for the text in Inkscape, but the package 'color.sty' is not loaded}%
    \renewcommand\color[2][]{}%
  }%
  \providecommand\transparent[1]{%
    \errmessage{(Inkscape) Transparency is used (non-zero) for the text in Inkscape, but the package 'transparent.sty' is not loaded}%
    \renewcommand\transparent[1]{}%
  }%
  \providecommand\rotatebox[2]{#2}%
  \newcommand*\fsize{\dimexpr\f@size pt\relax}%
  \newcommand*\lineheight[1]{\fontsize{\fsize}{#1\fsize}\selectfont}%
  \ifx\svgwidth\undefined%
    \setlength{\unitlength}{566.92913386bp}%
    \ifx\svgscale\undefined%
      \relax%
    \else%
      \setlength{\unitlength}{\unitlength * \real{\svgscale}}%
    \fi%
  \else%
    \setlength{\unitlength}{\svgwidth}%
  \fi%
  \global\let\svgwidth\undefined%
  \global\let\svgscale\undefined%
  \makeatother%
  \begin{picture}(1,1)%
    \lineheight{1}%
    \setlength\tabcolsep{0pt}%
    \put(0,0){\includegraphics[width=\unitlength,page=1]{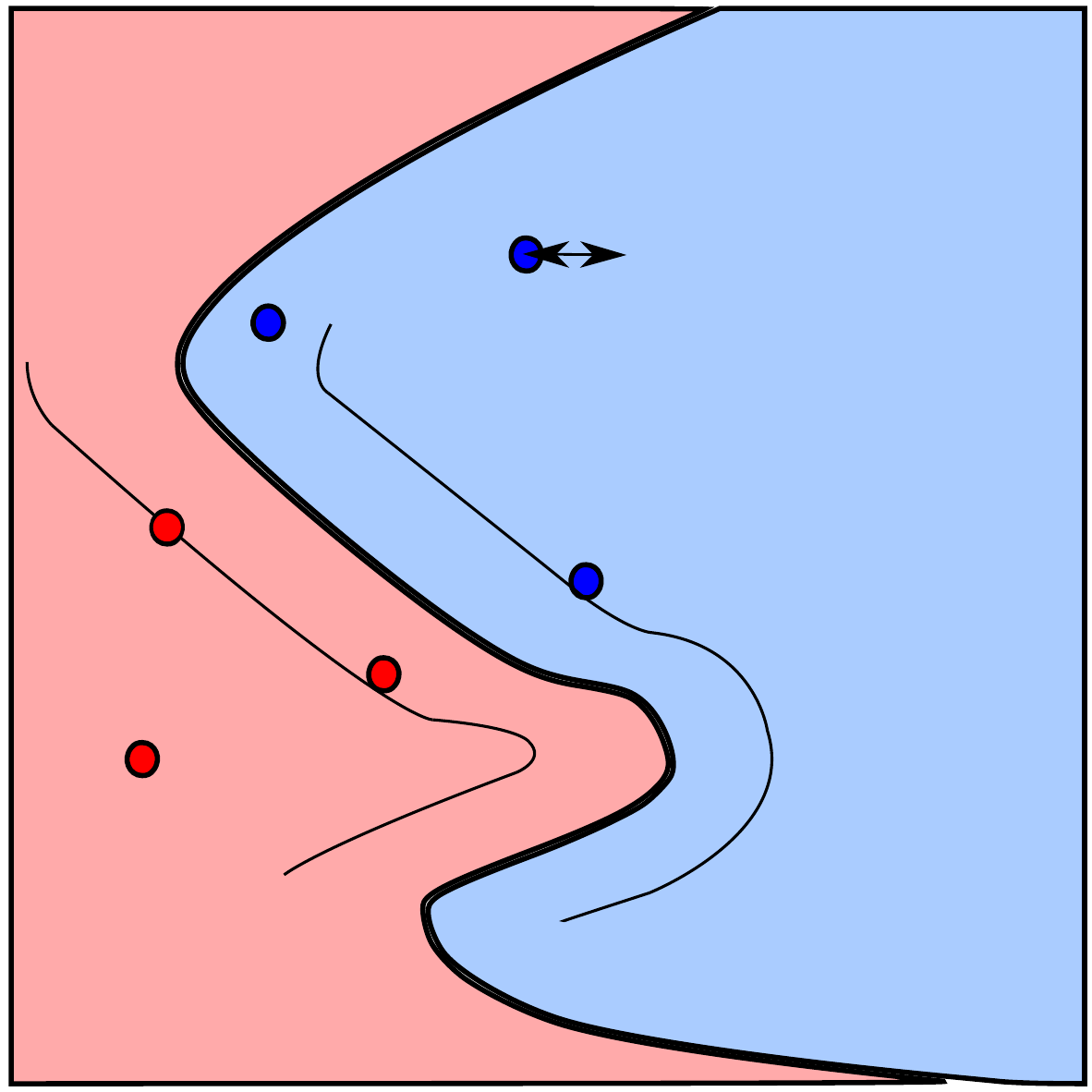}}%
    \put(0.5102769,0.71623892){\color[rgb]{0,0,0}\makebox(0,0)[lt]{\lineheight{1.25}\smash{\begin{tabular}[t]{l}$\delta$\end{tabular}}}}%
    \put(0,0){\includegraphics[width=\unitlength,page=2]{NhoodTrain2.pdf}}%
    \put(0.6542457,0.3080474){\color[rgb]{0,0,0}\makebox(0,0)[lt]{\lineheight{1.25}\smash{\begin{tabular}[t]{l}$\varepsilon$\end{tabular}}}}%
  \end{picture}%
\endgroup%

            \caption{Sample space of $\delta$-neighbourhoods about training points in which to estimate the $\text{\bf TrainBvol}$ $\varepsilon$-neighbourhood volume of the network decision boundary.}
            \label{fig:NhoodTrain}
        \end{figure}

        Providing $\delta$ is smaller than the minimal distance between points in $X$, the $\delta$-neighbourhoods around points do not overlap.
        In this case, uniformly sampling the disjoint union of the $B_{\delta}(x)$ for each $x\in X$ is equivalent to sampling the union.
        See Figure~\ref{fig:NhoodTrain} for a demonstration of the set $\mathcal{U}$ in this case.

        When $\delta$ is larger than the minimal distance between points $x\in X$ and points are sampled uniformly from the disjoint union of the $B_{\delta}(x)$, the probabilistic interpretation of the value approximated in equation~\eqref{eq:Bvol} becomes
        \begin{equation}\label{eq:NearProbInterpretation}
            \text{\bf TrainBvol} =
            |X| \cdot \mathbb{E}(\text{Bvol}_\varepsilon(N, B_\delta(x)) \; | \; x \in X).
        \end{equation}
        In practice, this is the measurement we make.
        However, it turns out that the theoretical and practical interpretations of \text{\bf TrainBvol} in equations~\eqref{eq:NearDef}~and~\eqref{eq:NearProbInterpretation} coincide for suitably high dimensional spaces.
        This equivalence is disused in Section~\ref{sec:CubeOverlap}.

    \subsection{\text{\bf LAdvBvol}}\label{sec:LAdvBvol}
        
        The final region $\mathcal{U}$ for neighbourhood boundary volume measurements is obtained in the following two steps.
        
\begin{figure}[ht!]
            \centering
            \def\svgwidth{200pt}
\begingroup%
  \makeatletter%
  \providecommand\color[2][]{%
    \errmessage{(Inkscape) Color is used for the text in Inkscape, but the package 'color.sty' is not loaded}%
    \renewcommand\color[2][]{}%
  }%
  \providecommand\transparent[1]{%
    \errmessage{(Inkscape) Transparency is used (non-zero) for the text in Inkscape, but the package 'transparent.sty' is not loaded}%
    \renewcommand\transparent[1]{}%
  }%
  \providecommand\rotatebox[2]{#2}%
  \newcommand*\fsize{\dimexpr\f@size pt\relax}%
  \newcommand*\lineheight[1]{\fontsize{\fsize}{#1\fsize}\selectfont}%
  \ifx\svgwidth\undefined%
    \setlength{\unitlength}{566.92913386bp}%
    \ifx\svgscale\undefined%
      \relax%
    \else%
      \setlength{\unitlength}{\unitlength * \real{\svgscale}}%
    \fi%
  \else%
    \setlength{\unitlength}{\svgwidth}%
  \fi%
  \global\let\svgwidth\undefined%
  \global\let\svgscale\undefined%
  \makeatother%
  \begin{picture}(1,1)%
    \lineheight{1}%
    \setlength\tabcolsep{0pt}%
    \put(0,0){\includegraphics[width=\unitlength,page=1]{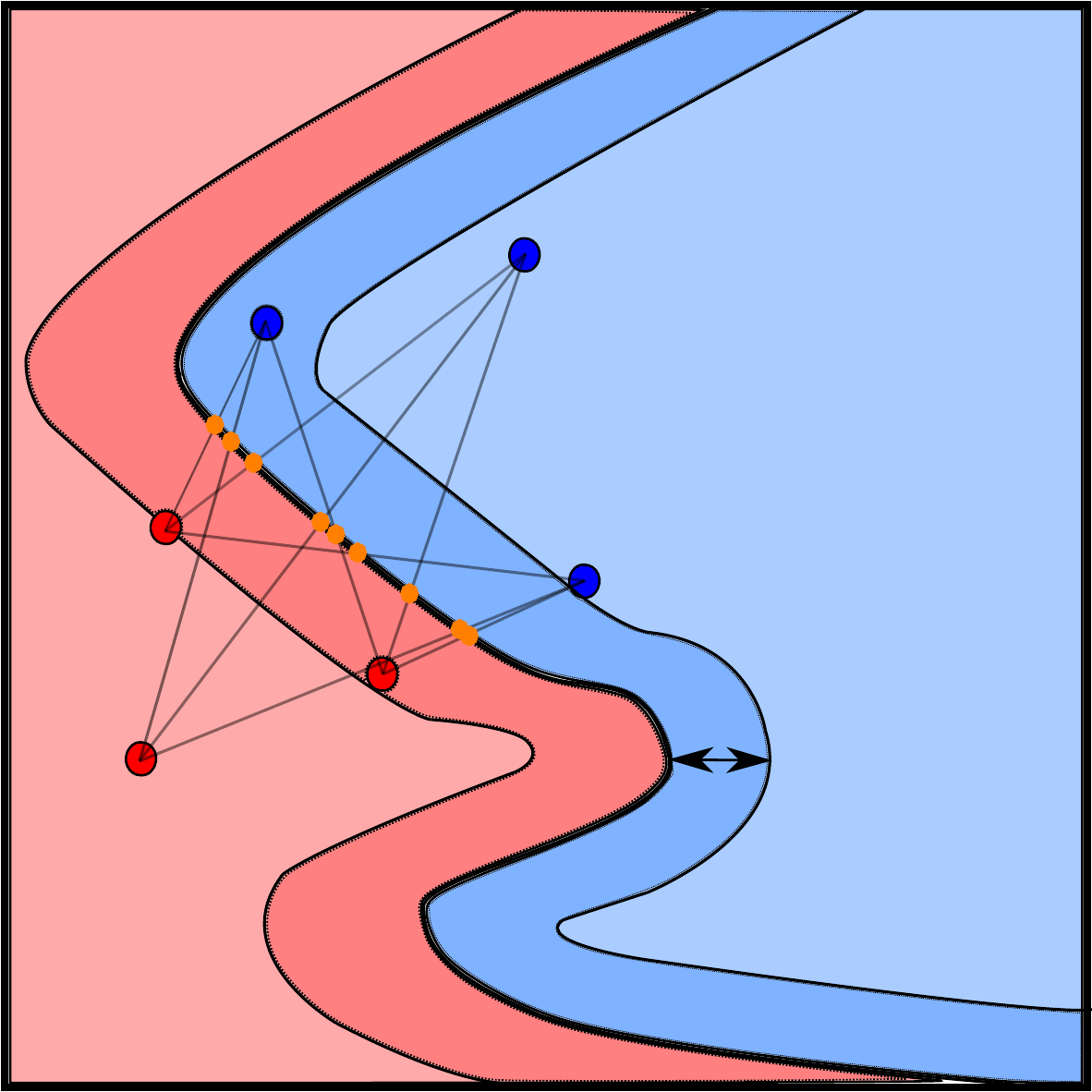}}%
    \put(0.65243295,0.32147155){\color[rgb]{0,0,0}\makebox(0,0)[lt]{\lineheight{1.25}\smash{\begin{tabular}[t]{l}$\varepsilon$\end{tabular}}}}%
    \put(0,0){\includegraphics[width=\unitlength,page=2]{NhoodBnd.pdf}}%
    \put(0.23422476,0.55348419){\color[rgb]{0,0,0}\makebox(0,0)[lt]{\lineheight{1.25}\smash{\begin{tabular}[t]{l}$\delta$\end{tabular}}}}%
  \end{picture}%
\endgroup%

            \caption{The points sampled on the boundary linearly between data classes in the first step (orange) within a $\delta$ neighbourhood of which the second step estimate of $\text{\bf LAdvBvol}$, the $\varepsilon$-neighbourhood volume of the network decision boundary
            is made.}
            \label{fig:NhoodBnd}
        \end{figure}

        Firstly, similarly to the procedures detailed in \cite{Guan2020}, we sample points lying linearly between two training examples of opposite classes in order to obtain a point on the decision boundary between them.
        These boundary points are given in orange in the demonstration provided in Figure~\ref{fig:NhoodBnd}.
        In practice, the boundary points are obtained by repeatedly selecting the midpoint of a line segment between points from opposite classes and determining a section of the line that must contain the boundary from the label of the midpoint.
        
        A bound on the proximity of these linear adversarial examples to the network decision boundary is determined by the number of line segment subdivisions $\alpha$.
        Assuming that the original two points are no more than distance $1$ apart, the distance of the linear adversarial example from the decision boundary is bounded above by $1/2^\alpha$.
        The output of the above procedure provides us with a sample of points located between label classes.
        
        Secondly, we measure how much the decision boundary twists between the data classes
        by making boundary volume measurements in $\delta$-neighbourhoods of the point sampled on the decision boundary above, similarly to the method from Section~\ref{sec:TrianBvol}.
        Again, see Figure~\ref{fig:NhoodBnd} for a demonstration of the procedure.

        Denote the set of pairs of points from $X$ with different class labels in $Y$ by
        \begin{align*}
            \text{cp}(X,Y) =
            \{ (x,x') \in X \times X \; | \; l(y)\neq l(y')) \}
        \end{align*}
        where $l(x) \in Y$ denotes the class label corresponding to $x\in X$.
        Define $LA(x,x')$ to be the closest adversarial example to $x$ lying on the line segment between points $x,x'\in \mathbb{R}^n$. 
        We define
        \begin{equation}\label{eq:BndDef}
            \text{\bf LAdvBvol} = \text{Bvol}_\varepsilon(N, \cup_{(x,x')\in \text{cp}(X,Y)}B_{\delta}(LA(x,x')).
        \end{equation}
        Similarly to the previous section, when the $\delta$-neighbourhoods are disjoint, the above expression coincides with what we measure in practice.
        For practical measurements in general, we consider the region
        \begin{equation*}
            \mathcal{U} = \coprod_{(x,x')\in \text{cp}(X,Y)} B_{\delta}(LA(x,x')),
        \end{equation*}
        in the case when the $B_{\delta}(LA(x,x'))$ do not overlap.
        In this case, the probabilistic interpretation of the $\varepsilon$-neighbourhood boundary volume measurement is
        \begin{align}\label{eq:BndProbInterpretation}
            \text{\bf LAdvBvol} =
            |\text{cp}(X,Y)| \cdot \mathbb{E}\big(&\text{Bvol}_\varepsilon(N, B_{\delta}(LA(x,x')) \; | \; (x,x') \in \text{cp}(X,Y)\big).
        \end{align}
        Again, the interpretations of \text{\bf TrainBvol} in equations~\eqref{eq:BndDef}~and~\eqref{eq:BndProbInterpretation} coincide for suitably high dimensional spaces,
        as will be detailed in Section~\ref{sec:CubeOverlap}.
        
        In addition, for practical use, even for a reasonably sized data set $X$ there is a very large number of possible pairs of points in $\text{cp}(X,Y)$.
        Therefore, in the experimental part of this work, we only take a random sample among the possible pairs $(x,x')$ with $l(x) \neq l(x')$ to obtain an estimate of the $\varepsilon$-neighbourhood boundary volume measurement $\text{\bf LAdvBvol}$.

    \subsection{Measuring decision boundaries in cubical regions}\label{sec:TubeInCube}

        Each of the $\varepsilon$-boundary volume measurements $\text{\bf Bvol}$, $\text{\bf TrainBvol}$, and $\text{\bf LAdvBvol}$ proposed in the previous subsections are confined to a region $\mathcal{U}$ consisting of a union of cubes in $\mathbb{R}^n$.
        The Weyl tube formula discussed in Section~\ref{sec:TubeFormula} defines the bounding region of a tubular neighbourhood to be determined by the direction of the normal vector on the manifold boundary.
        In low dimensions, the portion of the decision boundary that intersects a face of a cube need not have a normal vector lying inside that face.
        However, in high dimensions Proposition~\ref{prop:NAlmostOthonomal} states that randomly chosen vectors are almost certainly arbitrarily close to perpendicular.
        In particular, the normal vector to the section of network boundary intersecting a face of a cube is generically perpendicular to the normal vector of the face and therefore lies in the face. 
        This implies that the regions of the $\varepsilon$-neighbourhood of the decision boundary inside a cube or union of cubes are approximately the same as for Weyl's notion of a tubular neighbourhood.

    \subsection{Measuring $\text{\bf TrainBvol}$ and $\text{\bf LAdvBvol}$ in practice}\label{sec:CubeOverlap}

        The sample regions $\mathcal{U}$ considered when defining $\text{\bf TrainBvol}$ and $\text{\bf LAdvBvol}$ are formed in equations~\eqref{eq:NearDef}~and~\eqref{eq:BndDef} from the union rather than the disjoint union of a number of cubical regions centered on points lying in $[0,1]^n$.
        The disjoint union being the region we sample in practice, as described in equations~\eqref{eq:NearProbInterpretation}~and~\eqref{eq:BndProbInterpretation}.
        In particular, duplicate sampling of regions contained in multiple parts of the disjoint union changes the probabilistic interpretation of boundary volume measurements when compared to the original construction of $\text{Bvol}_\varepsilon(N,\mathcal{U})$ in Section~\ref{sec:BndVol} as a union of cubical regions.
        However, we justify here that in a high dimensional setting, under reasonable circumstances, it is unambiguous to refer to the practical measurement of $\text{\bf TrainBvol}$ and $\text{\bf LAdvBvol}$ as special cases of $\text{Bvol}_\varepsilon(N,\mathcal{U})$.

        More precisely, if two  unit cubes aligned with the coordinate axes in $\mathbb{R}^n$ overlap by $d_i$ in each coordinate axis for $i=1,\dots,n$, then the volume of the overlap region is $\prod_{i=1,\dots,n}d_i$. 
        Hence, given $m$ axis-aligned unit cubes in $\mathbb{R}^n$ of equal size with overlap in each axis bounded above by $0 \leq \zeta < 1$, we have
        \begin{equation}\label{eq:OverlapVolBound}
            \text{cube overlap volume } \leq \binom{m}{2}\zeta^n < \frac{m^2\zeta^n}{2}.
        \end{equation}
        Therefore, even if the number of cubes $m$ grows proportionally to the dimension $n$, the percentage volume of cube overlap vanishes as $n\to \infty$.
        Furthermore, in high dimensions the theoretical interpretations presented in equations~\eqref{eq:NearDef}~and~\eqref{eq:NearProbInterpretation} and the practical interpretations presented in equations~\eqref{eq:BndDef}~and~\eqref{eq:BndProbInterpretation} for \text{\bf LAdvBvol} and \text{\bf TrainBvol}
        converge to the same value as dimensions increase, respectively.
        
        As a demonstration, the value of the bound given in equation~\eqref{eq:OverlapVolBound} is plotted against the number of dimensions in the case of $100$ cubes for each dimension in Figure~\ref{fig:DimVsOverlap}.
        In the figure, we see that even for very large $\zeta$ values, the size of the bound reduces to almost zero in a relatively small number of dimensions.

    {\centering
    \begin{figure}[ht]
    		\begin{tabular}{cc}
    		    \begin{subfigure}{0.49\textwidth}\centering
    			\includegraphics[width=1\columnwidth]{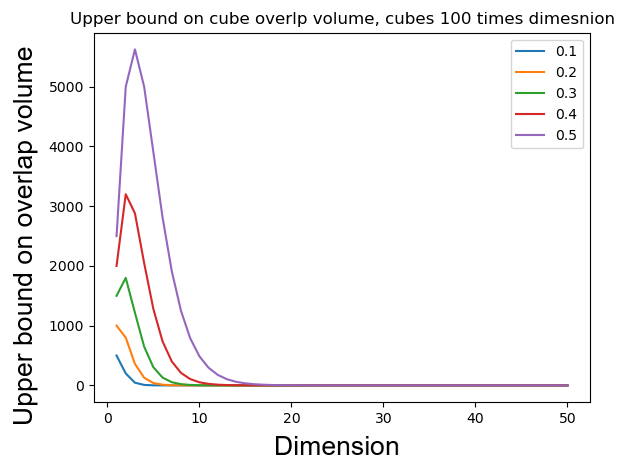}\caption*{}
    			\end{subfigure}&
    			\begin{subfigure}{0.54\textwidth}\centering
    			\includegraphics[width=1\columnwidth]{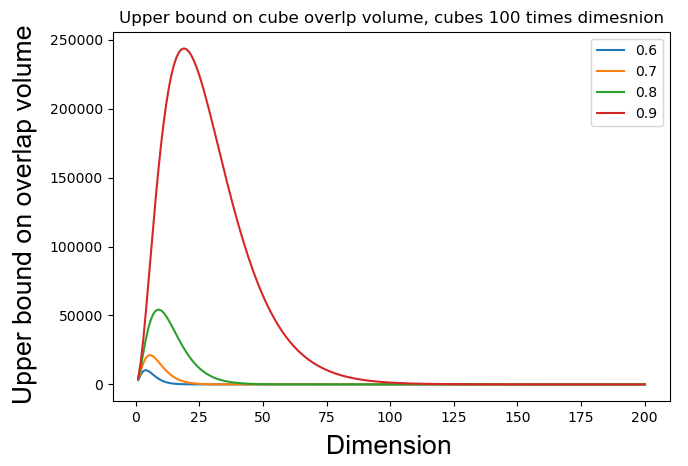}\caption*{}
    			\end{subfigure}
    		\end{tabular}
    	\caption{Value of the bound given in equation~\eqref{eq:OverlapVolBound} over varying dimensions in the case of $100$ cubes for each dimension, with curves for varying sizes of $\zeta$ between $0.1$ and $0.9$. The bound consistently vanishes for even modestly high dimensions.}
    	\label{fig:DimVsOverlap}
    \end{figure}}

\section{Experimental results}\label{sec:Experiments}

    In this section, we examine the relationship between decision boundary volume, overfitting, and generalisation of deep neural networks, demonstrating how the procedures set out in Sections~\ref{sec:methodology} can provide geometric explanations for network performance.
    We present experiments with neighbourhood boundary volume measurements on networks trained with fully connected and convolutional architectures under changes in dropout hyperparameter settings in their final hidden layer.
    In particular, our methods are applied to image data sets MNIST, Fashion MNIST, and CIFAR-10.
    The experiments presented in this section are intended as a proof of concept for the methods set out in this work. Therefore, we make use of data sets and network architectures for which boundary volume measurements can be computed easily using many Monte Carlo samples on multiple trained network functions.
    A TensorFlow implementation with examples of the procedures used in this work can be found in \cite{Burfitt2024}.

    \subsection{Summary of results}

    The reliability and error margin in the boundary volume measurements, in terms of $\varepsilon$ and $\delta$ values, and the number of Monte Carlo samples are considered alongside the selection of network parameter settings in Section~\ref{sec:ParameterChoices}.
    
    In Section~\ref{sec:Regularisation}, each data set and network architecture is trained with varying dropout rate
    between $p=0$ and $p=0.5$ at intervals of $0.1$ in their final hidden layer, in the manner introduced in Section~\ref{sec:Drop}.
    The neighbourhood boundary volumes $\textbf{TrainBvol}$ and $\textbf{LAdvBvol}$ are computed for each network architecture at $10$ random initialisations.
    
    For the convolutional neural networks, we observe a straightforward relationship between dropout rate and neighbourhood boundary volume.
    We see an increase in $\textbf{LAdvBvol}$ values as dropout rates increase and a local minimum in the $\textbf{TrainBvol}$ measurements coinciding with the dropout rate of highest test accuracy.
    For the fully connected networks, we observe a more complex trade-off between the $\textbf{LAdvBvol}$ measurements and the $\textbf{TrainBvol}$ measurements, 
    where the behavior of one measurement appears to drive the behavior of the other.
    More precisely, for any given data set a local maxima in the test accuracy correspond to a critical point in either the $\textbf{LAdvBvol}$ or $\textbf{TrainBvol}$ measurements.

    \subsection{Experimental setup}\label{sec:Setup}

    For the MNIST and Fashion MNIST data sets, fully connected networks with a single hidden layer of $100$ units were used. 
    On the CIFAR-10 data set, a fully connected network with $3$ hidden layers of $1024$ units was used.
    The architectures of the hidden layers of the convolutional neural networks are presented in Table~\ref{fig:ConvSetup}.
    All networks used a softmax activation on their output layer and rectified linear activation functions otherwise.

    \begin{table}[ht!]
      \begin{center}
        \begin{tabular}{c|c|c}
          MNIST & Fashion MNIST & CIFAR-10 \\
          \hline
           $\text{Conv } 16, \: 5\times 5+1$ & $\text{Conv } 16, \: 5\times 5+1$ & $\text{Conv } 16, \; 3\times 3+1$  \\
           $\text{Pool }2 \times 2+2\;\;\;\;\;$ & $\text{Pool }2 \times 2+2\;\;\;\;\;$ & $\text{Pool }2 \times 2+2\;\;\;\;\;$ \\
           $\text{Conv } 32, \: 5\times 5+1$ & $\text{Conv } 32, \: 5\times 5+1$ & $\text{Conv } 32, \: 3\times 3+1$ \\
           $\text{Pool }2 \times 2+2\;\;\;\;\;$ & $\text{Pool }2 \times 2+2\;\;\;\;\;$ & $\text{Pool }2 \times 2+2\;\;\;\;\;$ \\
           $\text{FC } 100\;\;\;\;\;\;\;\;\;\;\;$ & $\text{FC } 100\;\;\;\;\;\;\;\;\;\;\;$ & $\text{Conv } 64, \: 3\times 3+1$ \\
            &  & $\text{Pool }2 \times 2+2\;\;\;\;\;$ \\
            &  & $\text{FC } 256\;\;\;\;\;\;\;\;\;\;\;$
        \end{tabular}
      \end{center}
      \caption{Convolutional architectures used for training on data sets MNIST, Fashion MNIST, and CIFAR-10. The notation $\text{Conv } a, \; b\times b+1$ means a $2$-dimensional convolutional layer with $a$ square filters of size $b$, a stride of $1$, and same padding. All pooling layers $\text{Pool }2 \times 2+2$ used square filters of size $2$, a stride of $2$, and had no padding. The notation $\text{FC } c$ denotes a fully connected layer with $c$ hidden units.}
      \label{fig:ConvSetup}
    \end{table}

    The networks were trained with the cross-entropy loss function using randomly initialised weights distributed according to He normal conditions \cite{He2015}.
    The networks trained on MNIST with the stochastic gradient descent (SGD) optimiser used a learning rate of $0.01$. In all other cases, the networks trained on MNIST, Fashion MNIST, and CIFAR-10 with the Adam optimiser \cite{Kingma2015} used a learning rate of $0.0001$.
    In particular, each train iteration was completed with a constant batch size of $32$ in all cases.
    The number of training iterations varied between experiments and was chosen to coincide with the epoch at which the networks approximately first achieve their optimal training accuracy.
    We therefore detail the number of training epochs individually in the next section.
    
    In all experiments, the $\varepsilon$-neighbourhood boundary volume measurements used $\varepsilon$ and $\delta$ parameter settings detailed in Table \ref{fig:VolParameters}. 
    The Monte Carlo sample size in all cases was $10^5$, and the adversarial attack method was the FGSM.
    For \textbf{LAdvBvol} measurements, the number of subdivisions $\alpha$ used in the first step of the procedure was $10$, and the number of adversarial examples near the decision boundary sampled was $10^4$.

    \begin{table}[ht!]
      \begin{center}
        \begin{tabular}{c|c|c|c}
          Data & Method & $\varepsilon$ & $\delta$ \\
          (Nework) & & & \\
          \hline
                 & \textbf{Bvol}      & $0.001$ & - \\
           MNIST & \textbf{TrainBvol} & $0.003$  & $0.2$ \\
           (FC)  & \textbf{LAdvBvol}  & $0.001$  & $0.2$ \\
           \hline
           Fashion & \textbf{Bvol}    & $0.001$ & - \\
           MNIST & \textbf{TrainBvol} & $0.003$ & $0.2$ \\
           (FC)  & \textbf{LAdvBvol}  & $0.0008$ & $0.2$ \\
           \hline
                 & \textbf{Bvol}      & $0.002$ & - \\
           CIFAR-10&\textbf{TrainBvol}& $0.003$  & $0.2$ \\
           (FC)  & \textbf{LAdvBvol}  & $0.0008$ & $0.2$ \\
           \hline
                 & \textbf{Bvol}      & $0.0005$ & - \\
           MNIST & \textbf{TrainBvol} & $0.002$  & $0.2$ \\
           (Conv)& \textbf{LAdvBvol}  & $0.0003$ & $0.2$ \\
           \hline
           Fashion & \textbf{Bvol}    & $0.001$ & - \\
           MNIST & \textbf{TrainBvol} & $0.003$ & $0.2$ \\
           (Conv)& \textbf{LAdvBvol}  & $0.001$ & $0.2$ \\
           \hline
                 & \textbf{Bvol}      & $0.001$ & - \\
           CIFAR-10&\textbf{TrainBvol}& $0.0025$ & $0.05$ \\
           (Conv)& \textbf{LAdvBvol}  & $0.0005$ & $0.05$ \\
        \end{tabular}
      \end{center}
      \caption{The parameter values of $\varepsilon$ and $\delta$ for each neighbourhood boundary volume measurement using the FGSM introduced in Section~\ref{sec:FGSM}.}
      \label{fig:VolParameters}
    \end{table}

    \subsection{Justification for boundary volume parameter selection}\label{sec:ParameterChoices}

    The choices of $\delta$ in Table \ref{fig:VolParameters} were informed by considering the minimal distance between points in training set classes under the $d_\infty$ metric, based on the reasoning provided in Section~\ref{sec:CubeOverlap}.
    
    For MNIST, the minimal distance between two classes was approximately $0.737$ in the case of classes $1$ and $7$. With Fashion MNIST, all distances were above $0.4$ except for $5$ pairs of classes $(0,2),(1,6),(2,4),(2,6),(4,6)$. The shortest distance occurs in the $(2,4)$ case, at a value of $0.318$.
    Finally, for CIFAR-10, the minimal distance between classes was $0.22$.
    
    Based on the above distance measurements, for most experiments we chose a value of $\delta = 0.2$, as this was always less than the distance between classes.
    The only exception being the for training convolutional neural networks on CIFAR-10, where a smaller $\delta$ was necessary for clean results. In this case, $\delta=0.05$ was used.
    Therefore, applying the findings of Section~\ref{sec:CubeOverlap}, we conclude that in all cases $\textbf{TrainBvol}$ should coincide with the $\varepsilon$-neighbourhood boundary volume of the union of regions around the training data.
    Moreover, as can be seen from the findings presented in Table~\ref{fig:LinAdvPointDistance}, the above choices of $\delta$ values were also found to be reasonable for $\textbf{LAdvBvol}$ measurements. 
    This was despite some linear adversarial data points being at short $d_\infty$ distances from each other, as in general very few of the $10000$ sampled points lying close to the boundary were less than $0.2$ away from each other.
  
     \begin{table}[ht!]
      \begin{center}
        \begin{tabular}{c|c|c|c|c|c}
           Data & Minimal  & Average & Fraction & Fraction & Fraction \\
           (network) & distance  & distance & $< 0.2$ & $< 0.1$ & $< 0.05$ \\
          \hline
            MNIST (Fc)          & $0.0102$ & $0.587$ & $0.0034$ & $0.0008$ & $0.0004$ \\
            Fashion MNIST (Fc)  & $0.000400$ & $0.4601$ & $0.0269$ & $0.0181$ & $0.0145$ \\
            CIFAR-10 (Fc)       & $0.000270$ & $0.388$ & $0.0669$ & $0.039$ & $0.0313$ \\
            MNIST (Conv)        & $0.148$ & $0.577$ & $0.0002$ & $0$ & $0$ \\
            Fashion MNIST (Conv)& $0.000463$ & $0.465$ & $0.0215$ & $0.0057$ & $0.0026$ \\
            CIFAR-10 (Conv)     & $0.000216$ & $0.385$ & $0.0774$ & $0.0538$ & $0.0473$
        \end{tabular}
      \end{center}
      \caption{Statistics on the minimal $d_\infty$ distances between pairs of $10000$ linear adversarial points sampled using the method described in Sections~\ref{sec:LAdvBvol} for constructing points lying linearly between two training examples as part of the $\textbf{LAdvBvol}$ method. For networks trained with each combination of data sets and network architecture, the minimum distance and average minimum distances between pairs of the linear adversarial points are given alongside the fraction of those pairs of points that were less than $0.2$, $0.1$, or $0.05$ apart.}
      \label{fig:LinAdvPointDistance}
    \end{table} 
  
\begin{figure}
        \includegraphics[width=1\columnwidth]{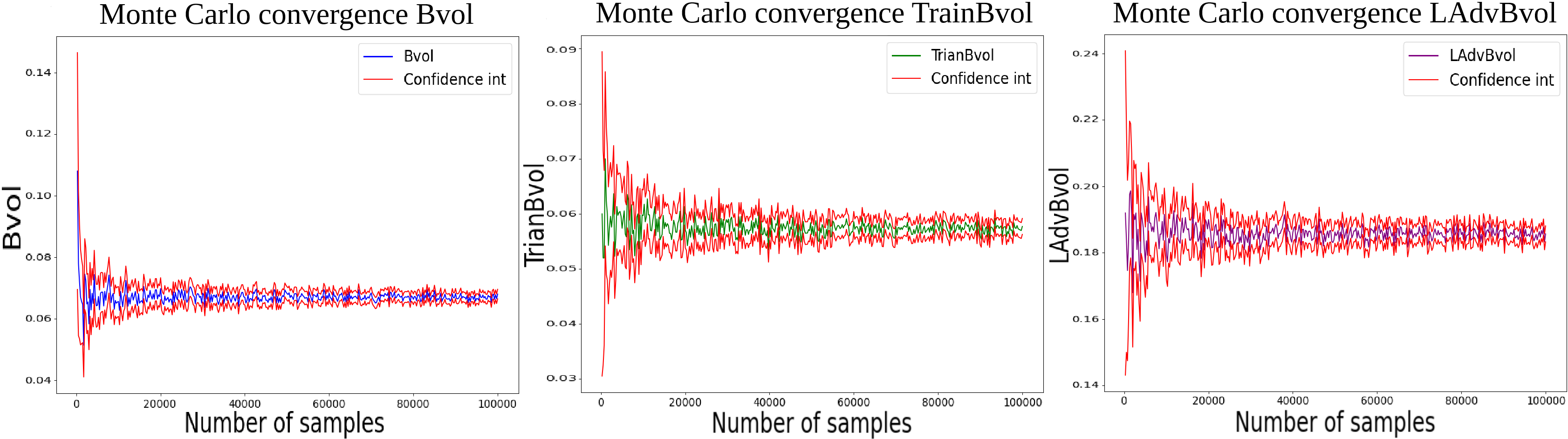}
        \caption{Values of $\textbf{Bvol}$, $\textbf{TrainBvol}$, and $\textbf{LAdvBvol}$ over a range of sample sizes plotted alongside $95\%$ confidence intervals obtained treating the Monte Carlo method as an estimation of a Bernoulli parameter (see Section~\ref{sec:BndVol}). The network used was fully connected and trained on MNIST with the Adam optimiser for $100$ epochs. In each plot the boundary volume measurements stabilise with reasonably sized confidence intervals when about $6000$ or more samples are taken.}
        \label{fig:MonteCarloConvergence}
    \end{figure}

    To test the size of the Monte Carlo error with $10^5$ point samples, we took a fully connected network trained on MNIST and plotted each of the boundary volume measurements alongside $95\%$ confidence intervals over a range of sample sizes. The results are presented in Figure~\ref{fig:MonteCarloConvergence}. In all cases the convergence has established a reasonable accuracy well before $10^5$ samples.
      
    Alternatively, evaluating the measure consecration bound given in equation~\eqref{eq:MCerrorBound2} under the scaling $\text{Volume}(\mathcal{U})=1$, we obtain that
    \begin{equation}\label{eq:EvaluatedBvolErrorBound}
        P\left( \left|\text{Bvol}_\varepsilon(N,\mathcal{U})-\widehat{\text{Bvol}}^{10^5}_\varepsilon(N,\mathcal{U})\right| \geq 0.01\right)  \leq 2.06  \times 10^{-9}
    \end{equation}
    where we recall that $\widehat{\text{Bvol}}^l_\varepsilon(N,\mathcal{U}))$ is the estimator of the actual value of $\text{Bvol}_\varepsilon(N,\mathcal{U})$ on $l$ samples.
    
    Given the findings presented above, the total number of Monte Carlo samples used to approximate $\textbf{TrainBvol}$ and $\textbf{LAdvBvol}$ was $10^5$. Both the number of training examples and the linear adversary sample size used for $\textbf{LAdvBvol}$ measurements were $10^4$.
    
        {\centering
        \begin{figure}[ht!]
        	\begin{tabular}{ccc}
        		\begin{subfigure}{0.33\textwidth}\centering
        			 \includegraphics[width=1\columnwidth]{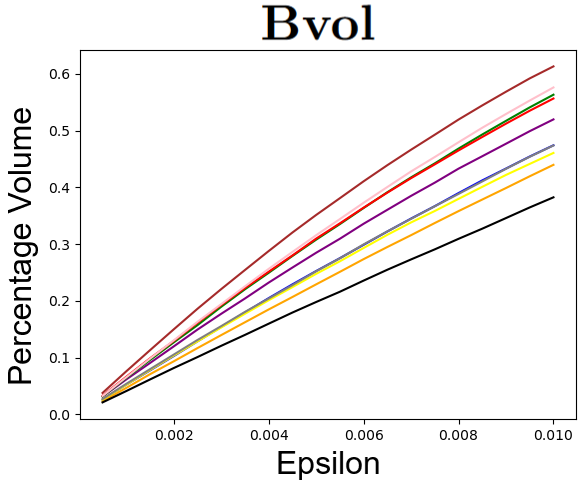}\caption*{}
        		\end{subfigure}&
        		\begin{subfigure}{0.33\textwidth}\centering
        			\includegraphics[width=1\columnwidth]{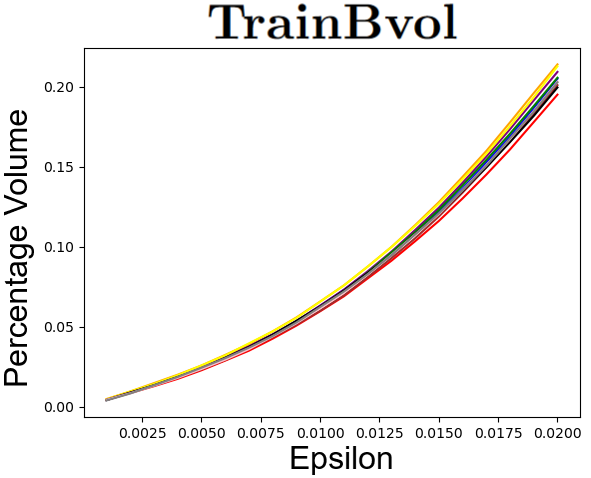}\caption*{}
        		\end{subfigure}
        		\newline
        		\begin{subfigure}{0.33\textwidth}\centering
        			\includegraphics[width=1\columnwidth]{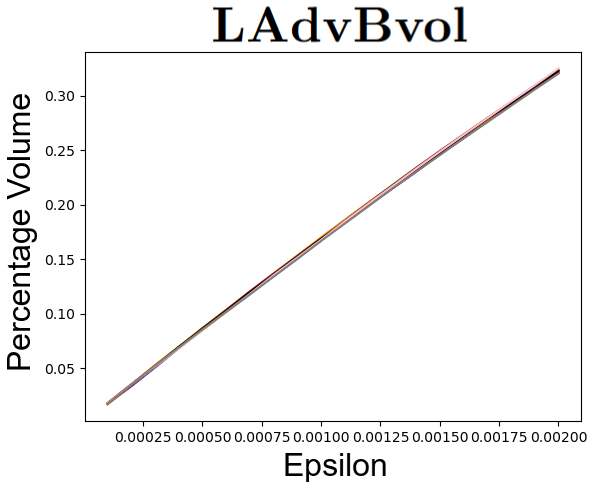}\caption*{}
        		\end{subfigure}
        	\end{tabular}
        	\caption{Values of $\varepsilon$-neighbourhood boundary volume measurements over a range of $\varepsilon$ values for $10$ networks trained on MNIST. The stability of the measurements appears good over the entire range of $\varepsilon$.
            We expect the rate of growth of the neighbourhood boundary volume measurements to decrease slightly for large values of $\varepsilon$, as the adversarial attack becomes less accurate at a greater distance. This feature is clearly apparent in the $\textbf{Bvol}$ and $\textbf{LAdvBvol}$ plots. However, in the $\textbf{TrainBvol}$ plot there is an initial increase in the rate of growth as $\varepsilon$ increases. Following the discussion in Sections~\ref{sec:TubeFormula}~and~\ref{sec:MeasuringBoundaries}, we would expect this to occur if curvature is present in this region of the decision boundary since the lower degree terms in Weyl’s tube formula are curvature dependent.}
        	\label{fig:10VolValRanges}
        \end{figure}}
    
    Lastly, we consider the selection of the $\varepsilon$ value.
    In Figure~\ref{fig:10VolValRanges}, the three $\varepsilon$-neighbourhood boundary volume measurements on $10$ networks trained with MNIST over a range of $\varepsilon$ values are plotted.
    In addition to remaining smooth over a range of values, each networks plot line is distinguishable, with measurement values keeping a consistent order. 
    This demonstrates that the choice of $\varepsilon$ value does not affect the measurements stability.

    Following the discussion in Section~\ref{sec:MeasuringBoundaries}, for a small enough $\varepsilon$, the $\varepsilon$-neighbourhood boundary volume measurements are approximately proportional to the true volume of the decision boundary we aim to measure.
    We make a choice of $\varepsilon$ values towards the lower end of the value ranges seen in Figure~\ref{fig:10VolValRanges} for the following reasons.
    
    Firstly, we chose $\varepsilon$ small enough to avoid the effects of curvature on measurements.
    That is, we chose $\varepsilon$ small enough so that as a function of the $\varepsilon$-neighbourhood boundary volume measurements in Figure~\ref{fig:10VolValRanges} are approximately locally linear.

    Secondly, it is necessary to select $\varepsilon$ values large enough so that the Monte Carlo errors do not become too high.
    In general, this can be tested in a similar way to the results presented in Figure~\ref{fig:MonteCarloConvergence} or using equation~\eqref{eq:EvaluatedBvolErrorBound}.

    \subsection{Effect of dropout on the decision boundary}\label{sec:Regularisation}

        We now investigate the effect of regularisation on the shape of the decision boundary of a trained neural network.
        In the experiments, we vary the values of dropout regularisation applied to the final hidden layers of the networks between $0$ and $0.5$ at intervals of $0.1$.
        In each case, we train $10$ randomly initialised networks presented in the columns of a graph in which the mean and one standard deviation on either side are also indicated.
        The number of training epochs selected for each experiment are given in Figure~\ref{fig:DropoutEpochs}. These values were chosen to be approximately the optimal training accuracy for each individual data set and dropout rate.  
        
        We present the boundary volume results for {\bf trainBvol} and {\bf LAdvBvol}, as together these contribute the most significant information.
        The {\bf Bvol} values
        are omitted as they provided less information due to measurement noise.
        
        First we provide the results on convolutional neural networks in Figure~\ref{fig:DropConv}.
        As a consequence of their more sophisticated architecture they provided a simpler structure in the behavior of their decision boundaries.
        After this we present the results on fully connected networks in Figure~\ref{fig:DropoutFc}.
        
        \begin{table}[ht!]
          \begin{center}
            \begin{tabular}{c|c|c|c|c}
                \textbf{Data} & \textbf{Architecture} & \textbf{Optimiser} & \textbf{Epochs} & \textbf{+epochs per $0.1$ dropout} \\
                \hline
                MNIST & Fc & SGD & 100 & 0 \\
                MNIST & Conv & SGD & 200 & 50 \\ 
                MNIST & Fc & Adam & 70 & 30  \\ 
                MNIST & Conv & Adam & 50 & 15 \\ 
                Fashion MNIST & Fc & Adam & 100 & 0 \\ 
                Fashion MNIST & Conv & Adam & 50 & 15 \\
                CIFAR-10 & Fc & Adam & 25 & 5 \\ 
                CIFAR-10 & Conv & Adam & 30 & 10
            \end{tabular}
          \end{center}
          \caption{Number of training epochs and increase in the number of training epochs per increase of $0.1$ of dropout rate on the final hidden layer. Epochs are given for fully connected and convolutional neural networks for each data set used in experiments in this section.}
          \label{fig:DropoutEpochs}
        \end{table} 
        
        In the first row of Figure~\ref{fig:DropConv}, we see the effect of varying the dropout rate on the final hidden layer of a convolutional neural network trained on MNIST with the SGD optimiser.
        In this case, the test accuracy continues to increase with higher dropout rates. The ${\bf TrainBvol}$ measurements decrease as the dropout rate is increased, even up to the point where the dropout rate becomes too high to stably train with, so the local minimum is the global minimum. 
        Meanwhile, the ${\bf {LAvBvol}}$ measurements increase monotonically as dropout rate increases.

        {\centering
        \begin{figure}[ht!]
        	\begin{tabular}{cccc}
        	    \begin{subfigure}{0.07\textwidth}\centering
                    
                \end{subfigure}&
        		\begin{subfigure}{0.3\textwidth}\centering
        			 Test accuracy
        		\end{subfigure}&
        		\begin{subfigure}{0.3\textwidth}\centering
        			TrainBvol
        		\end{subfigure}&
        		\begin{subfigure}{0.3\textwidth}\centering
        			LAdvBvol
        		\end{subfigure}\\
        		\begin{subfigure}{0.07\textwidth}\centering
        	        MNIST (SGD) \\ \:\\ \: \\ \:\\ \:
        		\end{subfigure}&
        		\begin{subfigure}{0.3\textwidth}\centering
        			 \includegraphics[width=1\columnwidth]{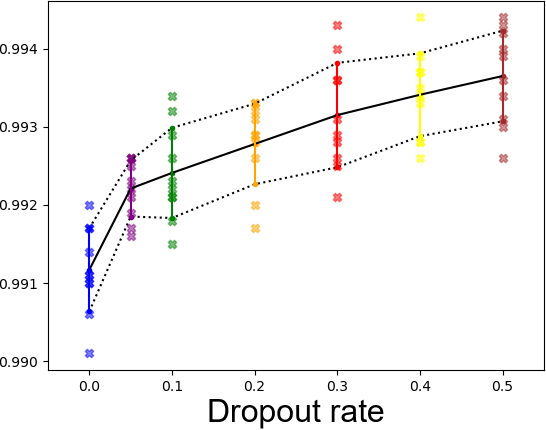}
        		\end{subfigure}&
        		\begin{subfigure}{0.3\textwidth}\centering
        			\includegraphics[width=1\columnwidth]{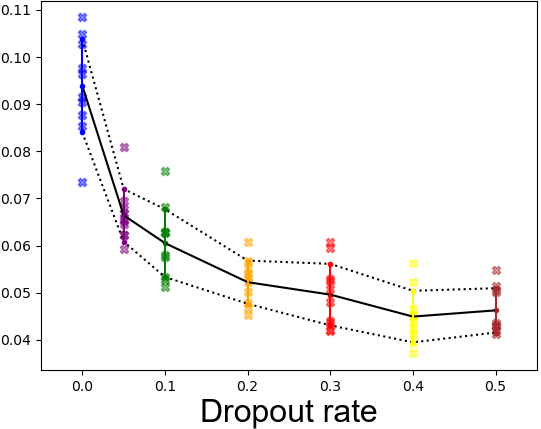}
        		\end{subfigure}&
        		\begin{subfigure}{0.3\textwidth}\centering
        			\includegraphics[width=1\columnwidth]{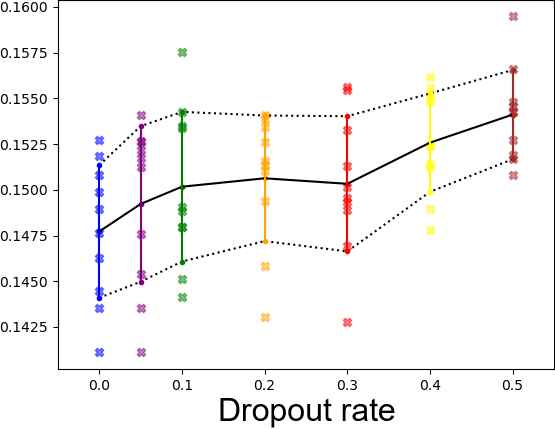}
        		\end{subfigure}\\
        		\begin{subfigure}{0.07\textwidth}\centering
        	        MNIST (Adam) \\ \:\\ \: \\ \:\\ \:
        		\end{subfigure}&
        		\begin{subfigure}{0.3\textwidth}\centering
        			 \includegraphics[width=1\columnwidth]{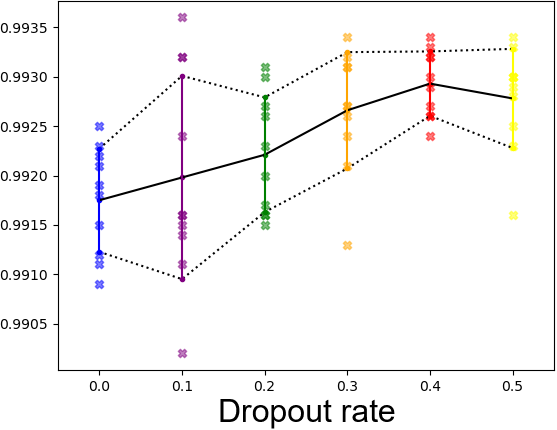}
        		\end{subfigure}&
        		\begin{subfigure}{0.3\textwidth}\centering
        			\includegraphics[width=1\columnwidth]{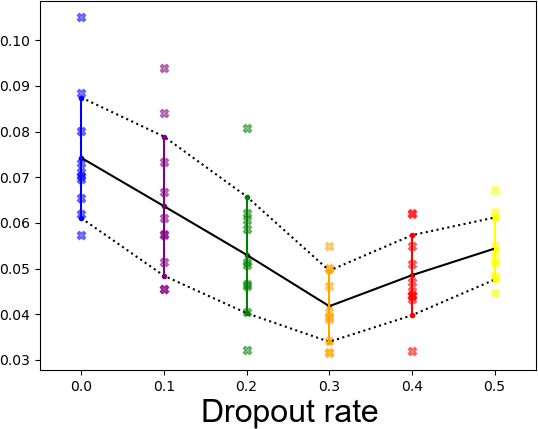}
        		\end{subfigure}&
        		\begin{subfigure}{0.3\textwidth}\centering
        			\includegraphics[width=1\columnwidth]{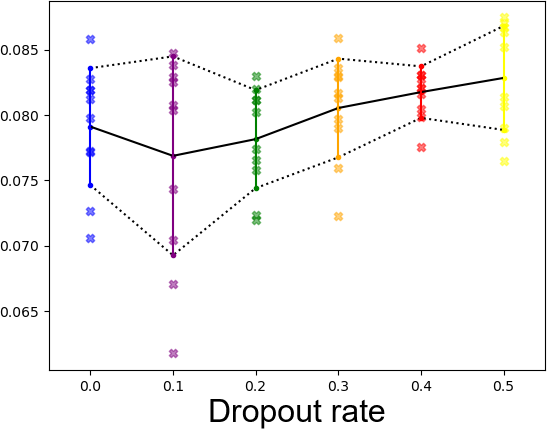}
        		\end{subfigure}\\
        		\begin{subfigure}{0.07\textwidth}\centering
        	        Fashion MNIST (Adam) \\ \:\\ \: \\ \:\\ \:
        		\end{subfigure}&
        		\begin{subfigure}{0.3\textwidth}\centering
        			 \includegraphics[width=1\columnwidth]{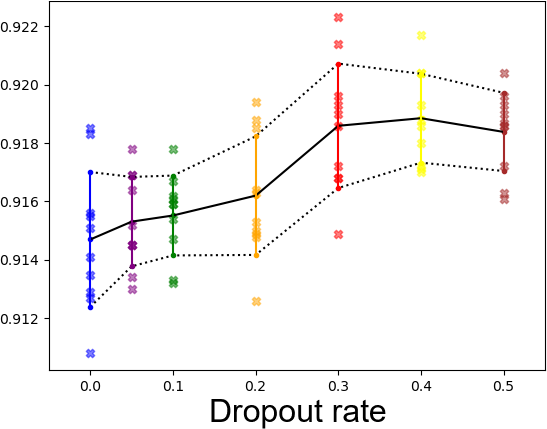}
        		\end{subfigure}&
        		\begin{subfigure}{0.3\textwidth}\centering
        			\includegraphics[width=1\columnwidth]{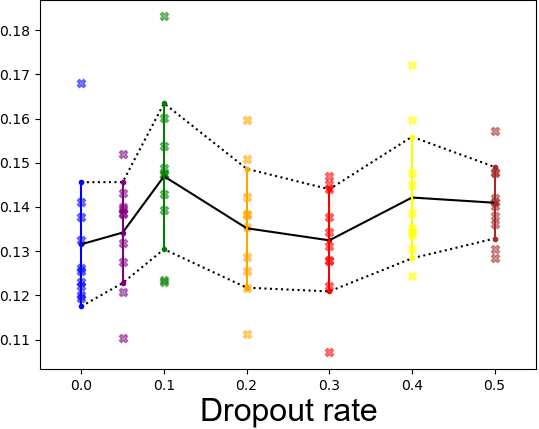}
        		\end{subfigure}&
        		\begin{subfigure}{0.3\textwidth}\centering
        			\includegraphics[width=1\columnwidth]{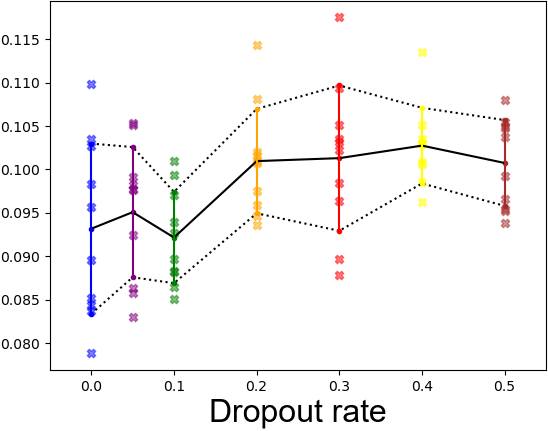}
        		\end{subfigure}\\
        		\begin{subfigure}{0.07\textwidth}\centering
        	        CIFAR-10 (Adam) \\ \:\\ \: \\ \:\\ \:
        		\end{subfigure}&
        		\begin{subfigure}{0.3\textwidth}\centering
        			 \includegraphics[width=1\columnwidth]{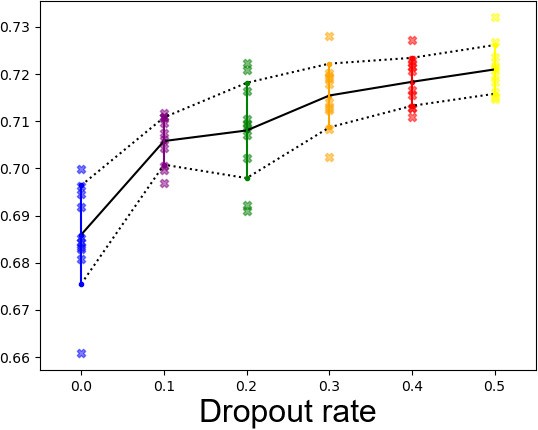}
        		\end{subfigure}&
        		\begin{subfigure}{0.3\textwidth}\centering
        			\includegraphics[width=1\columnwidth]{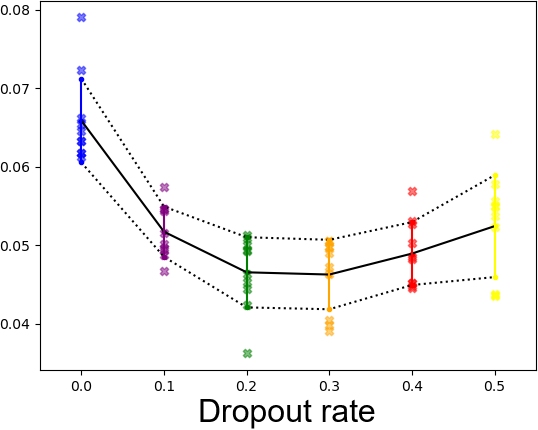}
        		\end{subfigure}&
        		\begin{subfigure}{0.3\textwidth}\centering
        			\includegraphics[width=1\columnwidth]{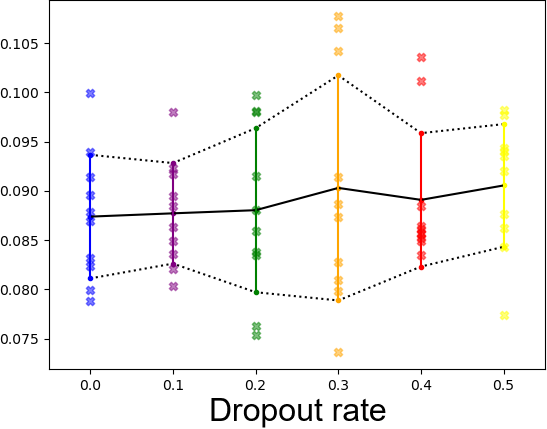}
        		\end{subfigure}
        	\end{tabular}
        	\caption{Boundary volume over varying dropout rate in the final fully connected layer applied to convolutional neural networks. The top two rows are trained on MNIST with the SGD and Adam optimisers, respectively. The third row is trained on Fashion MNIST, and the bottom row on CIFAR-10, both with the Adam optimiser. We see a local minima in $\textbf{TrainBvol}$ around or just before the optimal test accuracy, while the $\textbf{LAdvBvol}$ measurements generally increase.}
        	\label{fig:DropConv}
        \end{figure}}

        With the networks trained on MNIST with the Adam optimiser in the second row of Table \ref{fig:DropConv}, we see the test accuracy is highest around a dropout rate of $0.4$, lowering slightly on average by $0.5$. Similarly to the SGD case, the ${\bf {LAvBvol}}$ measurements generally increased monotonically with dropout rate, however, this time a minimum in the ${\bf TrainBvol}$ measurements appears at a dropout rate of $0.3$ just before the optimal dropout rate in the test accuracy. 
 
        For the networks trained on Fashion MNIST in the third row of Figure~\ref{fig:DropConv}, the test accuracy increases up to dropout rates of $0.3$ and $0.4$, before lowering slightly. The ${\bf TrainBvol}$ measurements decrease to a local minimum as dropout increases around $0.3$, corresponding to the best performing networks in terms of test accuracy. 
        The ${\bf {LAvBvol}}$ measurements vary little with a slight increase as dropout rate increases.
        
        Finally, in the last row of Figure~\ref{fig:DropConv}, the results for networks trained on CIFAR-10 show that the test accuracy increased with increased dropout rate, leveling off at rates between $0.3$ and $0.5$. As for the other data sets, the ${\bf {LAvBvol}}$ measurements can be seen to increase slightly with dropout rate, though in this case the results are considerably more noisy with a less steep slope.
        Meanwhile, the ${\bf TrainBvol}$ measurements fall to a local minima around dropout rates of $0.2$ and $0.3$ slightly preceding the leveling off in the test accuracy.
        
        {\centering
        \begin{figure}[ht!]
        	\begin{tabular}{cccc}
        	    \begin{subfigure}{0.07\textwidth}\centering
                    
                \end{subfigure}&
        		\begin{subfigure}{0.3\textwidth}\centering
        			 Test accuracy
        		\end{subfigure}&
        		\begin{subfigure}{0.3\textwidth}\centering
        			TrainBvol
        		\end{subfigure}&
        		\begin{subfigure}{0.3\textwidth}\centering
        			LAdvBvol
        		\end{subfigure}\\
        		\begin{subfigure}{0.07\textwidth}\centering
        	        MNIST (SGD)
        		\end{subfigure}&
        		\begin{subfigure}{0.3\textwidth}\centering
        			 \includegraphics[width=1\columnwidth]{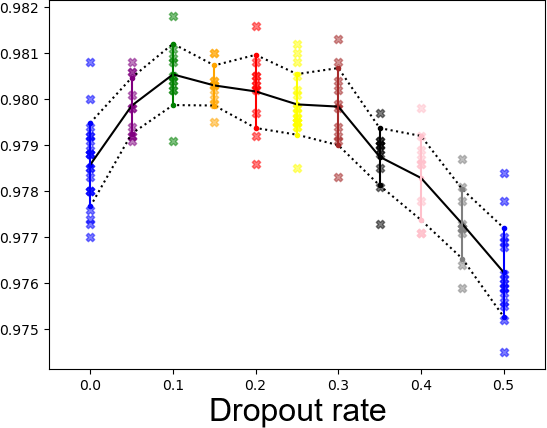}
        		\end{subfigure}&
        		\newline
        		\begin{subfigure}{0.3\textwidth}\centering
        			\includegraphics[width=1\columnwidth]{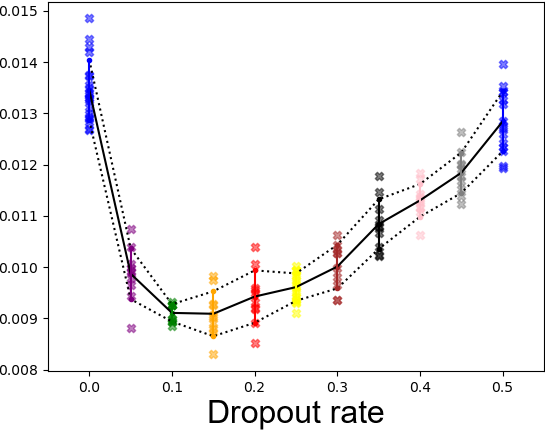}
        		\end{subfigure}&
        		\begin{subfigure}{0.3\textwidth}\centering
        			\includegraphics[width=1\columnwidth]{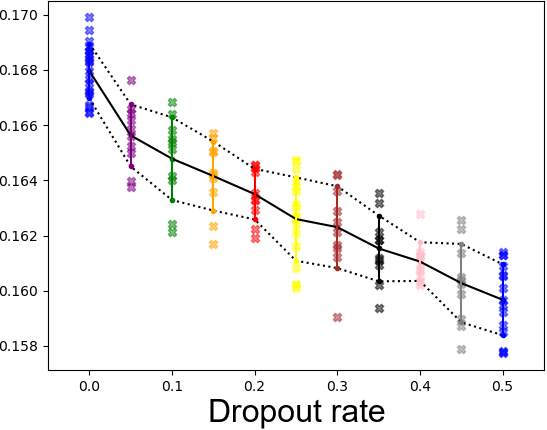}
        		\end{subfigure}\\
        		\begin{subfigure}{0.07\textwidth}\centering
        	        MNIST (Adam)
        		\end{subfigure}&
                \begin{subfigure}{0.3\textwidth}\centering
        			 \includegraphics[width=1\columnwidth]{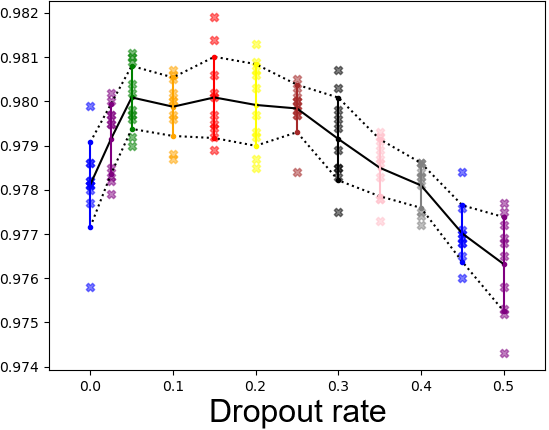}
        		\end{subfigure}&
        		\newline
        		\begin{subfigure}{0.3\textwidth}\centering
        			\includegraphics[width=1\columnwidth]{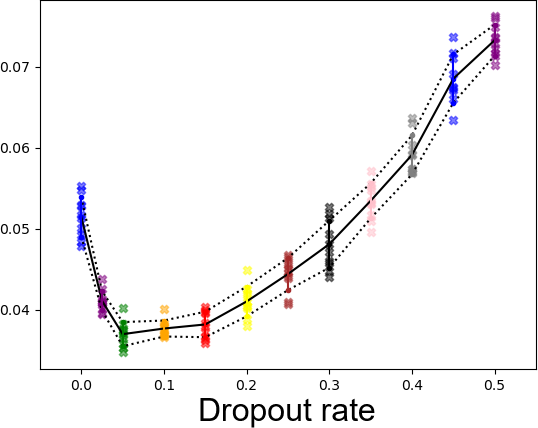}
        		\end{subfigure}&
        		\begin{subfigure}{0.3\textwidth}\centering
        			\includegraphics[width=1\columnwidth]{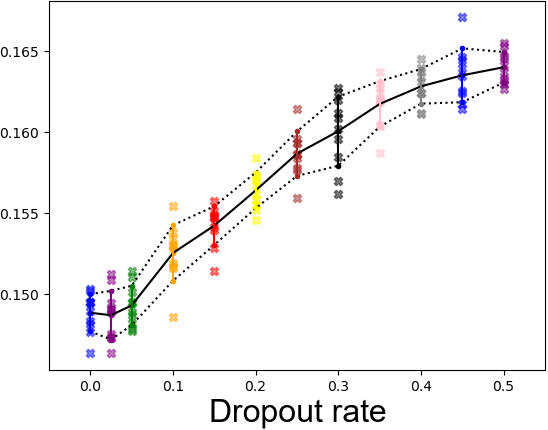}
        		\end{subfigure}\\
        		\begin{subfigure}{0.07\textwidth}\centering
        	        Fashion MNIST (Adam)
        		\end{subfigure}&
        		\begin{subfigure}{0.3\textwidth}\centering
        			 \includegraphics[width=1\columnwidth]{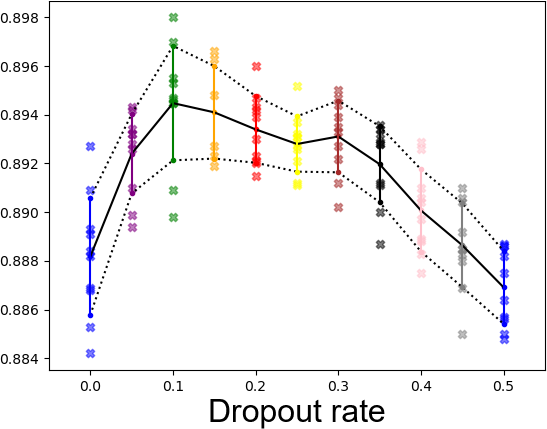}
        		\end{subfigure}&
        		\newline
        		\begin{subfigure}{0.3\textwidth}\centering
        			\includegraphics[width=1\columnwidth]{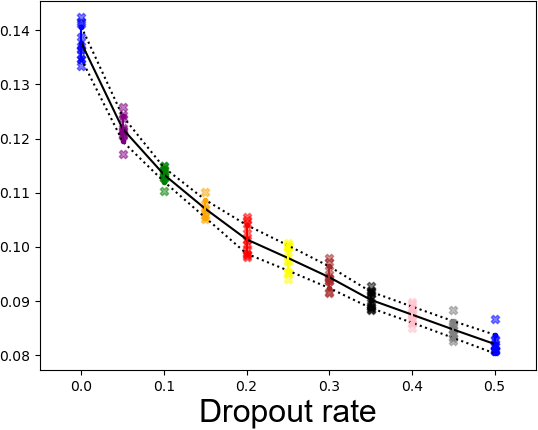}
        		\end{subfigure}&
        		\begin{subfigure}{0.3\textwidth}\centering
        			\includegraphics[width=1\columnwidth]{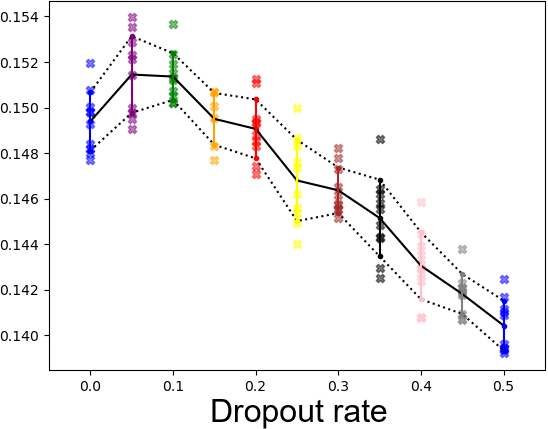}
        		\end{subfigure}\\
        		\begin{subfigure}{0.07\textwidth}\centering
        	        CIFAR-10 (Adam)
        		\end{subfigure}&
        		\begin{subfigure}{0.3\textwidth}\centering
        			 \includegraphics[width=1\columnwidth]{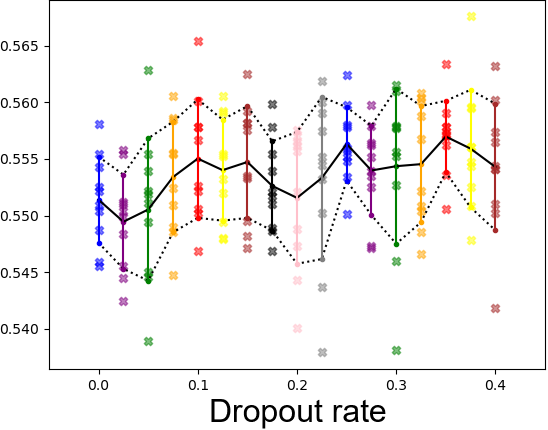}
        		\end{subfigure}&
        		\newline
        		\begin{subfigure}{0.3\textwidth}\centering
        			\includegraphics[width=1\columnwidth]{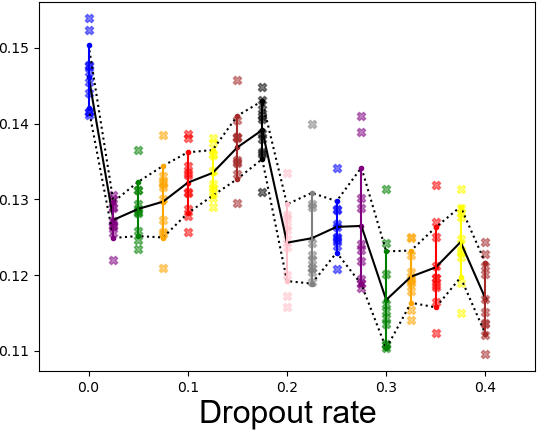}
        		\end{subfigure}&
        		\begin{subfigure}{0.3\textwidth}\centering
        			\includegraphics[width=1\columnwidth]{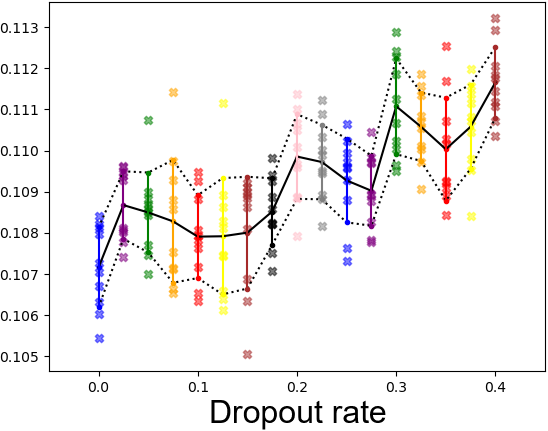}
        		\end{subfigure}
        	\end{tabular}
        	\caption{Boundary volume and test accuracy over varying dropout rates applied to fully connected neural networks. The top two rows are trained on MNIST with the SGD and Adam optimisers, respectively. The third row is trained on Fashion MNIST, and the bottom row is trained on CIFAR-10, both with the Adam optimiser. Optimal test accuracy is achieved around a critical point of either ${\bf TrainBvol}$ or ${\bf LAdvBvol}$, while the other of the two measurements changes monotonically.}
        	\label{fig:DropoutFc}
        \end{figure}}

        Figure~\ref{fig:DropoutFc} presents the values of the test accuracy and $\varepsilon$-neighbourhood boundary volume measurements for fully connected networks.
        The results for training on MNIST with the SGD and Adam optimisers over varying dropout rates in the first and second rows of Figure~\ref{fig:DropoutFc} show that on average the best test accuracies were obtained with a dropout rate around $0.1$.
        The ${\bf TrainBvol}$ measurements are minimised also around dropout values of $0.1$ and $0.15$, while the ${\bf LAdvBvol}$ values in the SGD case decrease monotonically and increase monotonically in the Adam case.
        
        An explanation for all the observations so far would be that increasing the dropout rate drives primarily a change in local behavior of ${\bf LAdvBvol}$, while there is then a trade off in the global behavior of ${\bf TrainBvol}$ values. It appears that the values at which this trade off finds network functions with better generalising properties are at the tuning point in ${\bf TrainBvol}$ measurements.
        However, it is surprising in the Adam case for fully connected networks on MNIST that increasing the regularisation hyperparameter initially leads to an increase in the complexity seen by both the ${\bf LAdvBvol}$ and ${\bf LAdvBvol}$ measurements.
        
        It should be noted that for results on MNIST, the ${\bf TrainBvol}$ measurements grow non-linearly in $\varepsilon$ (see Figure~\ref{fig:10VolValRanges}), which is not the case for other data sets. We expect this is due to curvature in the decision boundary and
        demonstrates why it was necessary to ensure that the values of $\varepsilon$ were sufficiently small.
        We observed that the effect of curvature decreased as the dropout rate increased as we would expect, this resulted in a shift to larger values of the ${\bf TrainBvol}$ local minima for larger $\varepsilon$ values.
        
        In the third row of Figure~\ref{fig:DropoutFc} are the results from the networks trained on Fashion MNIST with varying dropout rate.
        In this case ${\bf TrainBvol}$ decreases monotonically as the dropout rate increases, while ${\bf LAdvBvol}$ increases to a maximum before decreasing.
        This time the optimal test accuracy coincides with the change in behavior of the ${\bf LAdvBvol}$ measurements. So the roles of the ${\bf TrainBvol}$ and ${\bf LAdvBvol}$ measurements are reversed and the change in behaviour occurs around a local maximum rather than a local minimum. Suggesting a simplification in the complexity of the network function generally as the dropout rate is increased.

        The final row of Figure~\ref{fig:DropoutFc} contains the results from networks trained on CIFAR-10, and here we observe behavior similar to previous cases, but this time repeated multiple times in the boundary volume measurements over changes in dropout rate.
        Though noisy, there appear to be three local peaks in the average test accuracy occurring around dropout rates of $0.1$, $0.25$, and $0.35$, and these coincide with three local minima in the ${\bf LAdvBvol}$ plot.
        The values of ${\bf TrainBvol}$ also appear related to ${\bf LAdvBvol}$ measurements, as their behavior changes between increasing and decreasing shortly after the local minima in ${\bf LAdvBvol}$.
        In particular, increasing when ${\bf TrainBvol}$ decreases and decreasing when ${\bf TrainBvol}$ increases.

\section{Conclusion}
    In this paper we set out a method to measure the volume of a small neighbourhood of the decision boundary of a neural network. We identify three regions of the input space to measure, resulting in three neighbourhood boundary volume measurements $\text{\bf Bvol}$, $\text{\bf TrainBvol}$, and $\text{\bf LAdvBvol}$. Measuring these quantities allows us to deduce the properties of the network functions at varying scales and, hence, can be applied to provide a geometric explanation for properties of the network function.
    Using properties of concentration of measure in high dimensional spaces, we justified that for small neighbourhoods these neighbourhood boundary volume measurements are directly related to the volume of the decision boundary itself, up to multiplication by a scalar.
    Lastly, in the experimental part of our work, it is empirically demonstrated that boundary volume measurements can be efficiently computed for networks trained on large high dimensional data sets by applying a Monte Carlo method. 
    
    We apply our method to study generalisation in deep learning by measuring the decision boundaries of networks trained over changes in dropout regularisation hyperparameters. Here we observe that the generalisation properties of the network function are related to the structure of the decision boundary and that networks with better generalisation often occur at a critical point in the behavior of 
    the boundary volumes.
    This observation appears to be a consequence of a trade off between complexity at different scales, as demonstrated by the $\text{\bf TrainBvol}$ and $\text{\bf LAdvBvol}$, where one set of measurements demonstrates the critical behaviors while the other changes monotonically with respect to variations in the hyperparameter.
    For convolutional neural networks, the optimal generalisation consistently occurs around a local minimum of the $\text{\bf TrainBvol}$ neighbourhood boundary volume measurements taken within the vicinity of the training set.
    However, for fully connected neural networks, the nature of the connection between generalisation and decision boundary structures varies between data sets and even optimisation procedures, suggesting a more complex relationship between the two in general.

\bibliographystyle{amsplain}
\bibliography{Ref}

\appendix

\section{Proof of Theorem \ref{thm:RandomPlaneDistance}}\label{sec:Proof}

    Without loss of generality, we assume throughout the proof that $d_\infty(x,U) = 1$. 
    We now justifying the observation that quantities considered in parts (1) and (2) of Theorem~\ref{thm:RandomPlaneDistance} can be interpreted as the expectation and variance of the scalar product of the $n$-dimensional unit vector
    \[
        \hat{1} = \frac{1}{\sqrt{n}} (\underbrace{1,\dots,1}_{n})
    \]
    and a random unit vector in the region of $\mathbb{R}^n$ without negative coordinates, which we denoted by $\mathbb{R}^n_{\geq 0}$.
    The observation is provided by the following facts.
    \begin{enumerate}
        \item 
        Firstly, the problem is invariant under affine transformations, and hence it can be assumed that the hyperplane passes through the origin and that $x$ lies at $\hat{1}$.
        \item
        Secondly, a random hyperplane through the origin is entirely determined by its normal vector, and again applying (1), we can assume that the normal vector lies in $\mathbb{R}^n_{\leq 0}$.
        \item
        When $n\geq 3$, the affine transformation required to bring $x$ and the normal to the plane to $\hat{1}$ and $\mathbb{R}^n_{\geq 0}$, respectively, is not unique.
        However, all such transformations to any other choice of region between positive or negative coordinate axes are isometric to those that can be chosen in the case of $\mathbb{R}^n_{\geq 0}$.
    \end{enumerate}
    Therefore, after an affine transformation, we may identify the random unit vector $\hat{x}$ with the unit normal to the random hyperplane in the statement of the theorem and $\hat{1}$ with the point $x$ lying at a distance of $d_\infty(x,U) = 1$ away from the origin in the $d_{\infty}$ metric.
    The opposite direction of the normal vector $-\hat{x}$ being precisely the same as the direction that the point $\hat{1}$ should be projected onto the plane in order to obtain the point on the plane that is nearest to $\hat{1}$ with respect to the $d_2$ metric.   
    
    Finally, let $\theta$ be the angle between the normal to the hyperplane and $\hat{1}$.
    Using the geometric definition of a scalar product and the cosine rule, we have that
    \[
        \hat{x} \cdot \hat{1}
        =
        \cos{\theta}
        =
        \frac{d_2(\hat{1},U)}{d_\infty(\hat{1},U)}
    \]
    Applying this equality and part (3) above, we can now conclude that $\mathbb{E}(\hat{x} \cdot \hat{1}) = \mathbb{E}(d_2(x,U) / d_\infty(x,U))$.
    
    We now proceed to prove the statement of the theorem as translated by the observation above.
    However, we first introduced the necessary notions required for the proof.
    Some parts of the proof are a little harder to obtain in greater generality than necessary, so we include these as part of the argument.
    Moreover, deriving bounds for high order moments of $\hat{x}\cdot\hat{1}$ might in the future lead to better bounds similar to equation~\eqref{eq:ConsetrationOfNorm}, and the general statements we provide would be useful steps in this direction.
    
    Let $X$ be a non-negative random variable, then its central moment of order $k$ for $k=1,2,\dots$ is given by
    \[
     \mu_k = \mathbb{E}(|X-\mathbb{E}(X)|^k)
    \]
    which for the purposes of computation can be more easily expressed as
    \begin{equation}\label{eq:Moments}
        \mu_k = \sum_{i=0}^k (-1)^i \binom{k}{i}.
        \mathbb{E}(X)^i \mathbb{E}(X^{k-i})
    \end{equation}
    Suppose that $X$ has a central moment of order $k$.
    Applying Markov’s inequality to the random variable $|X-\mathbb{E}(X)|^k$, we obtain
    \[
        P(|X-\mathbb{E}(X)| \geq t) \leq
        \frac{\mathbb{E}(|X-\mathbb{E}(X)|^k)}{t^k}
    \]
    for each $t>0$.
    In the case when $k=2$, the above
    measure concentration bound is called Chebyshev’s inequality.
    Furthermore, once expressions for the $k$ order moments of $X$ less than some $k'$ are derived, we can  obtain $\mathbb{E}(X^{k'})$ using equation~\eqref{eq:Moments}. For this reason, we set out to derive expressions for the $k$ order moments $\mathbb{E}(|\hat{x}\cdot\hat{1}|^k)=\mathbb{E}((\hat{x}\cdot\hat{1})^k)$, which we achieve in full for $k=1$ and $k=2$.
    
    Let $\alpha$ denote the surface area of $S^{d-1}\cap\mathbb{R}^n_{\geq 0}$.
    Applying the multinomial theorem and changing to spherical coordinates, we see that
    \begin{align*}
        & \alpha \cdot \mathbb{E}((\hat{x}\cdot\hat{1})^k)
        \\ & =
        \frac{1}{n^{\frac{k}{2}}}
        \int_{x\in S^{n-1}\cap\mathbb{R}^n_{\geq 0}}
        (\hat{x}\cdot\hat{1})^k \:dA
        \\ & =
        \frac{1}{n^{\frac{k}{2}}}
        \int_{x\in S^{n-1}\cap\mathbb{R}^n_{\geq 0}}
        \left( \sum_{i=1}^n x_i \right)^k dA
        \\ & =
        \frac{1}{n^{\frac{k}{2}}}
        \int_{x\in S^{n-1}\cap\mathbb{R}^n_{\geq 0}}
        \sum_{k_1+\cdots+k_n=k}
        \binom{k}{k_1,\dots,k_n}
        x_1^{k_1}\cdots x_d^{k_n} \:dA
        \\ & =
        \frac{1}{n^{\frac{k}{2}}}
        \int^{\frac{\pi}{2}}_0\cdots\int^{\frac{\pi}{2}}_0
        \left( 
        \sum_{k_1+\cdots+k_n = k}
        \binom{k}{k_1,\dots,k_n}
        \prod_{i=1}^{n-1}\left(
        \cos^{k_i} \varphi_i
        \prod_{j=1}^{i-1} \sin^{k_i}\varphi_j
        \right)
        \prod_{l=1}^{n-1}
        \sin^{k_n}\varphi_l
        \right)
        \\ & \;\;\;\;\;\;\;\;\;\;\;\;\;\;\;\;\;\;\;\;\;\;\;\;\;\;\;\;
        \prod_{t=1}^{n-2} \sin^{n-t-1}\varphi_t
        \:d\varphi_1\cdots d\varphi_{n-1}
        \\ & =
        \frac{1}{n^{\frac{k}{2}}}
        \sum_{k_1+\cdots+k_n=k }
        \binom{k}{k_1,\dots,k_n}
        \int^{\frac{\pi}{2}}_0\cdots\int^{\frac{\pi}{2}}_0
        \prod_{i=1}^{n-1}
        \cos^{k_i}\varphi_{i}
        \sin^{n-i-1+k_{i+1}+\cdots+k_{n}}\varphi_i
        \:d\varphi_1\cdots d\varphi_{n-1}
        .
    \end{align*}
    For non-negative integers $a$ and $b$, it is well known that
    \begin{equation}\label{eq:TrigonometricIntegrals}
        \int^{\frac{\pi}{2}}_0\cos^{a}\theta\sin^b\theta \; d\theta
        = \frac{\Gamma(\frac{a+1}{2})\Gamma(\frac{b+1}{2})}{2\Gamma(\frac{a+b}{2}+1)}.
    \end{equation}
    Hence, we obtain
    \begin{align*}
        \alpha \cdot
        \mathbb{E}((\hat{x}\cdot\hat{1})^k)
        = &
        \frac{1}{n^{\frac{k}{2}}}
        \sum_{k_1+\cdots+k_n=k }
        \binom{k}{k_1,\dots,k_n}
        \prod_{i=1}^{n-1}
        \frac{\Gamma\left(\frac{k_i+1}{2}\right)\Gamma\left(\frac{n-i+k_{i+1}+\cdots+k_{n}}{2}\right)}{2\Gamma\left(\frac{n-i-1+k_{i}+\cdots+k_{n}}{2}+1\right)}
        \\ = &
        \frac{1}{2^{n-1}n^{\frac{k}{2}}}
        \sum_{k_1+\cdots+k_n=k }
        \binom{k}{k_1,\dots,k_n}
        \prod_{i=1}^{n-1}
        \frac{\Gamma\left(\frac{k_i+1}{2}\right)\Gamma\left(\frac{n-i+k_{i+1}+\cdots+k_{n}}{2}\right)}{\Gamma\left(\frac{n-i+1+k_{i}+\cdots+k_{n}}{2}\right)}
        \\ = &
        \frac{1}{2^{n-1}n^{\frac{k}{2}}}
        \sum_{k_1+\cdots+k_n=k }
        \binom{k}{k_1,\dots,k_n}
        \frac{\Gamma\left(\frac{1+k_{n}}{2}\right)}{\Gamma\left(\frac{n+k_{1}+\cdots+k_{n}}{2}\right)}
        \prod_{i=1}^{n-1}
        \Gamma\left(\frac{k_i+1}{2}\right)
        \\ = &
        \frac{1}{2^{n-1}n^{\frac{k}{2}}}
        \sum_{k_1+\cdots+k_n=k }
        \binom{k}{k_1,\dots,k_n}
        \frac{\prod_{i=1}^{n}
        \Gamma\left(\frac{k_i+1}{2}\right)}{\Gamma\left(\frac{n+k}{2}\right)}
        \\ = &
        \frac{1}{2^{n-1}n^{\frac{k}{2}}
        \Gamma\left(\frac{n+k}{2}\right)}
        \sum_{k_1+\cdots+k_n=k }
        \binom{k}{k_1,\dots,k_n}
        \prod_{i=1}^{n}
        \Gamma\left(\frac{k_i+1}{2}\right)
        .
    \end{align*}
    For the case $k=1$, in each summand above, all the $k_i$ for $i=1,\dots,n$ are zero except for some $j=1,\dots,n$ where $k_j = 1$.
    In particular, the multinomial coefficients all have value $1$, and the $\Gamma\left( \frac{k_i+1}{2} \right)$ are all equal to $\sqrt{\pi}$ except when $i=j$, in which case it is $1$.
    Therefore,
    \begin{align}\label{eq:GeneralExpression}
        \alpha \cdot \mathbb{E}(\hat{x}\cdot\hat{1})
        = &
        \frac{\pi^{\frac{n-1}{2}}\sqrt{n}}{2^{n-1}\Gamma\left( \frac{n+1}{2} \right)}
        .
    \end{align}
    It can be deduced form the surface area of $S^{d-1}$, that
    \begin{equation}\label{eq:SphereRegionVol}
        \alpha
        =
        \frac{\pi^{\frac{n}{2}}}{2^{n-1}\Gamma\left( \frac{n}{2} \right)}
            .
    \end{equation}
    Using the final expression of the calculation above, setting $k=1$, and dividing by $\alpha$, we arrive at
    \begin{equation}\label{eq:ExpectedValue}
        \mathbb{E}(\hat{x}\cdot\hat{1})
        =
        \frac{\sqrt{n}}{\sqrt{\pi}}
        \frac{\Gamma(\frac{n}{2})}{\Gamma(\frac{n+1}{2})}
        .
    \end{equation}
    To deduce the variance of $\hat{x}\cdot\hat{1}$, we first set $k=2$ in equation~\eqref{eq:GeneralExpression}.
    However, now there are two distinct possible configurations appearing in the summands.
    Either some $k_j=2$ and all other $k_i=0$ or $k_{j_1}= k_{k_2}=1$ for some $1 \leq j_1 < j_2 \leq n$ and all other $k_i=0$.
    In particular, in the first case the multinomial coefficients all have value $1$ and the $\Gamma\left( \frac{k_i+1}{2} \right)$ are all equal to $\sqrt{\pi}$ except when $i=j$ with $\Gamma\left( \frac{k_j+1}{2} \right) = \frac{\sqrt{\pi}}{2}$.
    For the second case, the multinomial coefficients all have value $2$, and the $\Gamma\left( \frac{k_i+1}{2} \right)$ are all equal to $\sqrt{\pi}$ except when $i=j_1$ or $i=j_2$, for which they have value $1$.
    Therefore, we have
    \begin{align*}
        \mathbb{E}\left((\hat{x}\cdot\hat{1})^2\right)
        = &
        \frac{1}{2^{n-1}n\Gamma\left( \frac{n+2}{2} \right)}
        \left(
        n\frac{\pi^{\frac{n}{2}}}{2}
        +
        2\frac{n(n-1)}{2}\pi^{\frac{n-2}{2}}
        \right)
        \\ = &
        \frac{\pi^{\frac{n}{2}}}{2^{n}\Gamma\left( \frac{n}{2} + 1 \right)}
        \left(
        1
        +
        \frac{2}{\pi}(n-1)
        \right).
    \end{align*}
    Again using equation~\eqref{eq:SphereRegionVol}, dividing by $\alpha$, and the fact that for $z\in\mathbb{C}$ with non-negative real part $\Gamma(z+1) = z \Gamma(z)$, we arrive at
    \begin{equation*}
        \mathbb{E}\left((\hat{x}\cdot\hat{1})^2\right)
        =
        \frac{\Gamma\left( \frac{n}{2} \right)}{2\Gamma\left( \frac{n}{2} + 1 \right)}
        \left(\frac{2}{\pi}(n-1) + 1\right)
        =
        \frac{1}{n}\left(\frac{2}{\pi}(n-1) + 1\right)
        =
        \frac{2}{\pi}\frac{n-1}{n}+\frac{1}{n}.
    \end{equation*}
    Moreover, using Gautschi's inequality \cite{Gautschi1959}, 
    we have
    \begin{equation*}
        \frac{\Gamma\left(\frac{n+1}{2}\right)}{\Gamma\left(\frac{n}{2}\right)}
        = 
        \frac{\Gamma\left(\frac{n-1}{2}+1\right)}{\Gamma\left(\frac{n-1}{2}+\frac{1}{2}\right)}
        > 
        \left(\frac{n}{2}\right)^{\frac{1}{2}}.
    \end{equation*}
    Therefore, using the formula for the expected value from equation~\eqref{eq:ExpectedValue} and the inequality above, we obtain the variance bound
    \begin{align*}
        \text{Var}(\hat{x}\cdot\hat{1})
        & =
        \mathbb{E}\left((\hat{x}\cdot\hat{1})^2\right)
        -
        \left(\mathbb{E}(\hat{x}\cdot\hat{1})\right)^2
        \\ & =
        \frac{2}{\pi}\frac{n-1}{n}+\frac{1}{n}
        -
        \frac{n}{\pi}
        \left(
        \frac{\Gamma(\frac{n}{2})}{\Gamma(\frac{n+1}{2})}
        \right)^2
        \\ & \leq
        \frac{2}{\pi}\frac{n-1}{n}+\frac{1}{n}
        -\frac{2}{\pi}
        \\ & =
        \frac{\pi - 2}{n}
        .
    \end{align*}
    This completes the proof of Theorem~\ref{thm:RandomPlaneDistance}.

\end{document}